\documentclass[lettersize,journal]{IEEEtran}
\usepackage{amsmath,amsfonts}
\usepackage{algorithmic}
\usepackage{algorithm}
\usepackage{array}
\usepackage[caption=false,font=normalsize,labelfont=sf,textfont=sf]{subfig}
\usepackage{textcomp}
\usepackage{stfloats}
\usepackage{url}
\usepackage{verbatim}
\usepackage{graphicx}
\usepackage{cite}

\makeatletter
\let\NAT@parse\undefined
\makeatother
\usepackage[colorlinks,citecolor=green]{hyperref}

\usepackage{epstopdf}
\usepackage{makecell}
\usepackage{xcolor}
\usepackage{tabularx}
\usepackage{supertabular}
\usepackage{pdflscape}
\usepackage{caption}
\usepackage{multirow}

\begin{document}

\title{From Attributes to Natural Language: A Survey and Foresight on Text-based Person Re-identification}

\author{

        \IEEEauthorblockN{
        Fanzhi Jiang, 
        Su Yang\IEEEauthorrefmark{1}, 
        Mark W. Jones, 
        Liumei Zhang\IEEEauthorrefmark{1}
        }
        
        \thanks{\IEEEauthorrefmark{1}Corresponding author}
\thanks{This paper was produced by the IEEE Publication Technology Group. They are in Piscataway, NJ.}
\thanks{Manuscript received April 20, 2024; revised August 16, 2024.}
        }

\markboth{Journal of \LaTeX\ Class Files,~Vol.~14, No.~8, December~2023}%
{Shell \MakeLowercase{\textit{et al.}}: A Sample Article Using IEEEtran.cls for IEEE Journals}

\maketitle

\begin{abstract}
  Text-based person re-identification (Re-ID) is a challenging topic in the field of complex multimodal analysis, its ultimate aim is to recognize specific pedestrians by scrutinizing attributes/natural language descriptions. Despite the wide range of applicable areas such as security surveillance, video retrieval, person tracking, and social media analytics, there is a notable absence of comprehensive reviews dedicated to summarizing the text-based person Re-ID from a technical perspective. To address this gap, we propose to introduce a taxonomy spanning $Evaluation$, $Strategy$, $Architecture$, and $Optimization$ dimensions, providing a comprehensive survey of the text-based person Re-ID task. We start by laying the groundwork for text-based person Re-ID, elucidating fundamental concepts related to attribute/natural language-based identification. Then a thorough examination of existing benchmark datasets and metrics is presented. Subsequently, we further delve into prevalent feature extraction strategies employed in text-based person Re-ID research, followed by a concise summary of common network architectures within the domain. Prevalent loss functions utilized for model optimization and modality alignment in text-based person Re-ID are also scrutinized. To conclude, we offer a concise summary of our findings, pinpointing challenges in text-based person Re-ID. In response to these challenges, we outline potential avenues for future open-set text-based person Re-ID and present a baseline architecture for text-based pedestrian image generation-guided re-identification $(TBPGR)$.
\end{abstract}

\begin{IEEEkeywords}
Person Re-Identification, Text, Natural Language, Attributes, Diffusion Model.
\end{IEEEkeywords}

\section{Introduction}
\begin{figure*}[!t]
  \centering
  \includegraphics*[width=\textwidth]{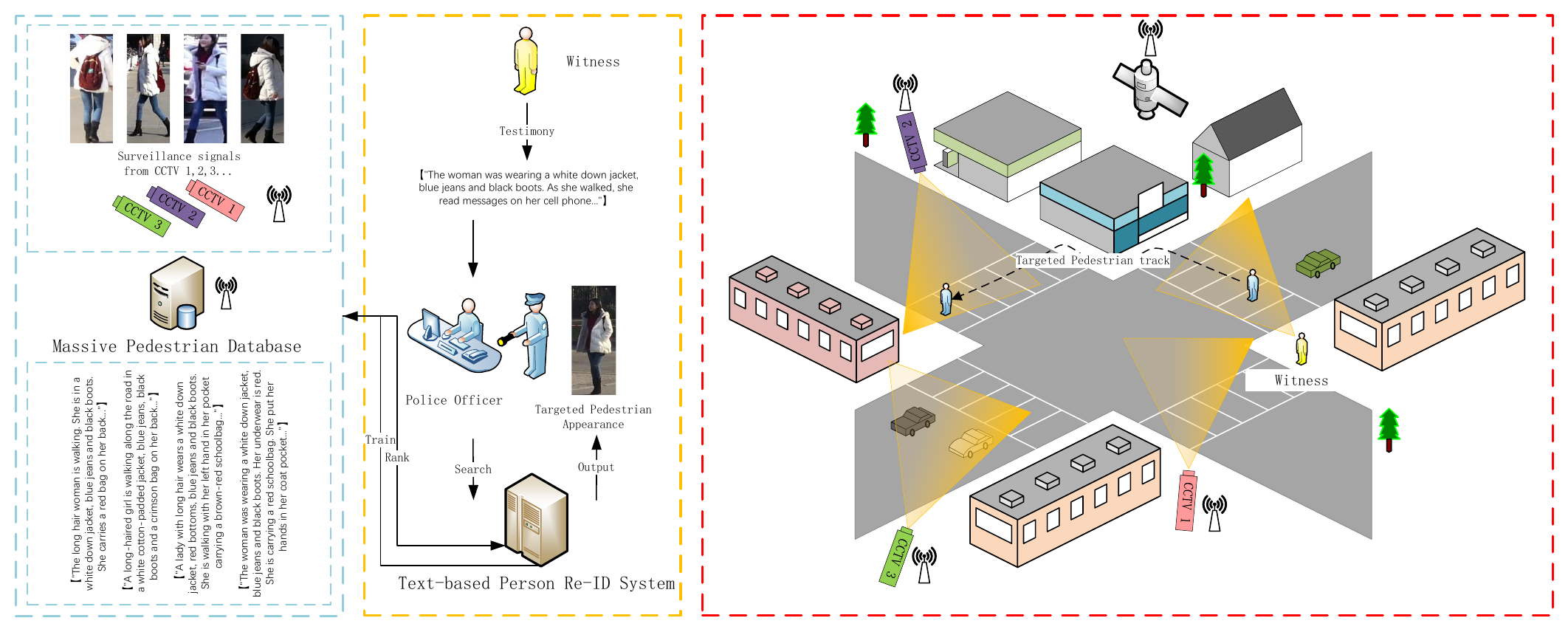}
  \caption{Conceptual diagram of text-based person Re-ID. Given a textual description of a target person collected from a street witness, a monitor uses the model aimed at retrieving the corresponding person image from a given database of images collected from street CCTV.}
  \label{fig1}
\end{figure*}

\IEEEPARstart{T}{HE} demands in public safety and the subsequently installed surveillance networks make the manual tracking and identifying individuals increasingly challenging. 
Automatic person re-identification (Re-ID) has emerged as a popular research area in computer vision to address this issue. Also known as person search ~\cite{yeDeepLearningPerson2022}, person Re-ID specifically refers to recognizing and tracking specific individuals using non-overlapping images and videos captured by cameras, determining whether a specific query person appears persistently or momentarily across different times and locations ~\cite{caiParallelDataAugmentation2022a}. 
It finds important applications in scenarios like searching for characters in movies, locating missing children, and tracking criminals ~\cite{wangIntramodalitySurveyHeterogeneous2020}. 
The source data collection for these applications is usually from CCTV footage on the street. Given the constraints of the equipment and the environment, a number of challenges arise. 
These challenges include lighting variations, occlusions, background changes, pose variations, and differences in camera resolutions. 
A typical person re-identification system involves conducting image queries and searching for corresponding individuals in a gallery of images or a video pool. 
It involves extracting features directly from pedestrian appearance images and performing direct matching, then ranking against existing appearance images in the gallery. 
These scenarios assume that visual examples of individual identities are always available as queries. However, some special cases where the visualization samples of personal identity are not available, we perform retrieval based on textual descriptions only, which is called text-based person re-identification, as shown in Figure \ref{fig1}.

\subsection{Attribute-based Person Re-ID}
Text-based person re-identification is a special form of person re-identification that, instead of using image data, relies on descriptions to reflect a person's appearance ~\cite{liPersonSearchNatural2017}. In the task of text-based person Re-ID, a natural language description and an image are provided as input, and the output is the identification of the person matching the description ~\cite{chenTIPCBSimpleEffective2022}. There has been a desire to perform high-level semantic person search using free-form natural language descriptions.  However, early research initially started with the attribute-based person re-identification \cite{hirzerPersonReidentificationDescriptive2011,xiongPersonReIdentificationUsing2014,linImprovingPersonReidentification2019,dengPedestrianAttributeRecognition2014a}. 
Attribute-based methods (Pedestrian Attributes Recognition, PAR) stem from the high correlation between attributes and pedestrian images. Pedestrian attributes such as gender, age, clothing type, clothing color, etc. are arranged and combined to make the pedestrian identity appear distinguishable \cite{laynePersonReidentificationAttributes2012}.
Extracting attribute information from pedestrian images can be achieved by using pre-trained classifiers or attribute detectors. These classifiers or detectors can recognize and locate different attributes(body part) in the images. These extracted pedestrian attributes information needs to be encoded into a feature vectors pattern matching. This often involves transforming discrete attributes into continuous feature representations, common methods include one-hot encoding and embedding encoding \cite{dengPedestrianAttributeRecognition2014a}. During the recognition process, matching is conducted by comparing the attribute feature vectors of query images and gallery images. Common matching methods include computing distances or similarities between feature vectors (such as Euclidean distance, cosine similarity, etc.). Finally, based on the results of attribute matching, the final recognition result is determined through fusion of matching scores from different attributes or by adopting decision strategies.
The attribute-based person Re-ID approach improves recognition accuracy and retrieval interpretability. This is because these attributes can be considered as high-level semantic information that is robust to viewpoint changes and different viewing conditions \cite{wangPedestrianAttributeRecognition2019}.

Previous work on attribute-based person retrieval has attempted to reduce modality gaps by aligning each person category and its corresponding images in a joint embedding space through modal adversarial training \cite{caoSymbioticAdversarialLearning2020, yinAdversarialAttributeimagePerson2018}, or by enhancing the expressive power of embedding vectors for person categories and images hierarchically \cite{dengPedestrianAttributeRecognition2014a}. While these pioneering studies reveal important but less explored methods for person retrieval, there remains significant room for further improvement. Firstly, due to their adversarial learning strategy, they exhibit instability and high computational cost during training \cite{caiParallelDataAugmentation2022a,jingPoseGuidedMultiGranularityAttention2020a}. For other, they incur high inference costs due to the need for additional networks to match high-dimensional embedding vectors \cite{dengPedestrianAttributeRecognition2014a}. More importantly, these methods treat person categories as independent labels of person images and overlook their relationships, such as how many attributes differ between them, despite the potential rich supervisory signals for learning better representations of person categories and images. Dong et al. ~\cite{dongPersonSearchText2019} address this issue by capturing rich information of two modalities through hierarchical embedding. However, their model entails high computational complexity as it computes high-dimensional embeddings and deploys additional networks for matching. Yin et al. ~\cite{yinAdversarialAttributeimagePerson2018} and Cai et al. ~\cite{caiParallelDataAugmentation2022a} learn a joint embedding space where person categories and images are directly matched. To bridge the modality gap, their embedding space is trained in a modal adversarial manner; however, due to the nature of min-max optimization, this often leads to unstable and slow convergence. Furthermore, all these methods suffer from a limitation where person categories are treated as separate labels and the important relationships among them are overlooked. In the joint embedding space of the two modalities, the loss pulls images of the same person category closer to achieve modal alignment. \cite{jeongASMRLearningAttributeBased2021} introduces a novel loss, Adaptive Semantic Margin Regularizer (ASMR), for learning cross-modal embeddings in the context of attribute-based person retrieval.

\begin{figure} 
  \centering
  \includegraphics[width=3.0in]{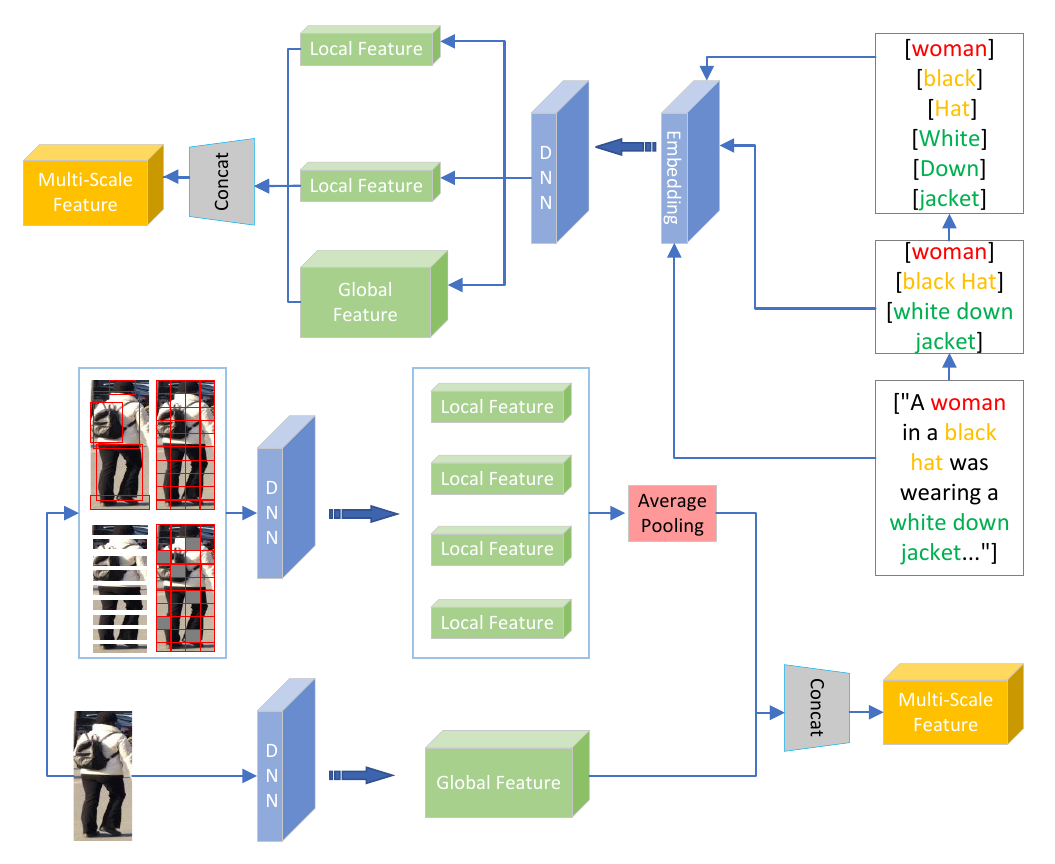}
  \caption{Multi-scale feature fusion: The figure shows the basic operations regarding the fusion of image and text multi-scale features in person re-identification.}
  \label{fig3}
\end{figure}

\subsection{Natural Language-based Person Re-ID}
Attribute-based person retrieval \cite{zhengHierarchicalGumbelAttention2020,liIdentityAwareTextualVisualMatching2017} provides a method for person re-identification based on free-form descriptions. 
However, there is an awareness that the leverage of attributes to describe a person's appearance could be limited in some circumstances. For instance, the PETA dataset \cite{dengPedestrianAttributeRecognition2014a} defines 61 binary and 4 multi-class person attributes, along with hundreds of phrases used to describe a person's appearance.  In addition, even with a constrained set of attributes, labeling them for large-scale person image datasets is challenging and expensive. Using free-form natural language (NL) descriptions may offer more flexibility and robustness compared to these predefined attributes. 

Attribute-based person retrieval also has limitations in terms of the results robustness. For instance, shuffling and flipping some attributes while using a basic fact set for retrieval can lead to significant performance degradation of the results. Compared to manually assigned sets of attributes, textual descriptions can be seen as an unstructured form of annotation. In general, most attributes fall into the category of noun phrases, and natural language includes higher-level semantic words and complex events than the attributes. In principle, natural language processing (NLP) techniques can facilitate all attributes with any possible words and lengths, capturing unique details through natural language descriptions. Additionally, descriptions can include indicators of certainty or fuzziness to retain as much information as possible, potentially allow sets of multi-class attribute assignments. Both attribute-based and NLP-based person re-identification are referred to as text-based person re-identification in this paper.

Li et al. \cite{linPersonSearchChallenges2021a} introduced the first dataset for NL-based person Re-ID, evaluated and compared various possible models, and proposed an RNN with a gated neural attention mechanism. Jing et al. ~\cite{jingPoseGuidedJointGlobal2018} proposed to focus on mining fine-grained local parts and performing fine-grained visual-textual matching. These methods can learn distinctive part representations and achieve 81.23\% accuracy, but they are troubled by the problem of ambiguous embeddings. Image regions and words are correlated at different semantic granularities \cite{wangViTAAVisualTextualAttributes2020a,zhengHierarchicalGumbelAttention2020}. The discriminative parts of a person usually appear with different granularities, which was overlooked in the previous methods and ambiguated embeddings. To alleviate this issue, words (based on NL) as guidance have been integrated into the related image regions. However, irrelevant words may mislead the model and even exacerbate the ambiguous embeddings problem. 

Cross-domain embeddings technique is one of the main trends in the Re-ID community. Work proposed in \cite{chenImprovingDeepVisual2018} added supplementary annotations to extract meaningful image regions independently, these regions, however, are not always available in the real-world scenes. Work in \cite{liIdentityAwareTextualVisualMatching2017} proposed to learn cross-modal embeddings to improve matching results using a shared attention mechanism. \cite{zhangDeepCrossModalProjection2018a} designed a cross-modal projection matching loss and a cross-modal projection classification loss for learning discriminative image-text embeddings. \cite{yanCrossdomainPersonReidentification2022b} proposed a matrix factorization-based dictionary learning algorithm to eliminate the impact of style and pedestrian pose information on cross-domain Re-ID. \cite{chenImprovingTextBasedPerson2018} proposed a patch-word matching model and designed an adaptive threshold mechanism in the model, it attempted to establish global and local image-language associations to enhance the semantic consistency between local visual and language features. 

Deep adversarial graph attention convolutional network was also proposed to address the problem of text-based person re-identification\cite{sarafianosAdversarialRepresentationLearning2019,liuDeepAdversarialGraph2019}. Niu et al. \cite{niuImprovingDescriptionBasedPerson2020}proposed a Multi-level Image-Text Alignment (MIA) model, which introduced cross-modal adaptive attention in three different granularities, integrating these three similarities to comprehensively judge the affinity of image-sentence pairs and better distinguish interested pedestrians. Zhu et al. \cite{zhuDSSLDeepSurroundingsperson2021}achieved an environment-person separation (DSSL) method under the mutual exclusion constraint, and also annotated another dataset for person search using language called Real Scene Text-based Person Re-ID (RSTPReID). Wu et al. \cite{wuLapsCoreLanguageGuidedPerson2021} designed a twin task, image coloring and text completion, and explicitly built fine-grained cross-modal associations bidirectionally based on color reasoning, their LapsCore method achieved 63.4\% rank@1 retrieval accuracy on the CUHK-PEDES dataset. \cite{niuCrossmodalCooccurrenceAttributes2022} proposed a novel cross-modal matching method named Cross-modal Co-occurrent Attribute Alignment (C2A2), which can better handle noise and significantly improve retrieval performance in person search by language. Relying on matrix factorization to construct visual and textual attribute dictionaries and cross-modal alignment using denoising reconstruction features addresses noise generated by extraneous elements of pedestrians. Zheng et al. \cite{zhengDualpathConvolutionalImageText2020} made the first attempt to perform NL-based person search within unsegmented complete images.

\subsection{Person Re-Identification previous survey}
In order to provide researchers with a comprehensive understanding of the current status and research directions in the domain of text-based person re-identification, we conducted an in-depth investigation of text-based person re-identification methodologies and synthesized recent research achievements. Prior to this effort, several researchers have conducted reviews in the field of person re-identification \cite{wangIntramodalitySurveyHeterogeneous2020,bedagkar-galaSurveyApproachesTrends2014b,zhengPersonReidentificationMeets2015,chaharStudyDeepConvolutional2017,wangSurveyPersonReidentification2018,wuOverviewDeepLearning2022, mathurBriefSurveyDeep2020, islamPersonSearchNew2020,laviSurveyReliableDeep2020,zouPersonReidentificationBased2021,yangSurveyUnsupervisedTechniques2021a,yaghoubiSSSPRShortSurvey2021,wangCrossDomainPersonReidentification2021,mingDeepLearningbasedPerson2022,pengDeepLearningBasedOccluded2023,vinyalsShowTellNeural2015, zahraPersonReidentificationRetrospective2023, singhComprehensiveSurveyPerson2022,huangDeepLearningVisibleinfrared2023} and we summarize the primary contributions of these studies in the Table \ref{table1}. 
Among them, certain surveys \cite{bedagkar-galaSurveyApproachesTrends2014b,chaharStudyDeepConvolutional2017} introduced previous challenges in the field of person Re-ID. \cite{wangSurveyPersonReidentification2018,wuOverviewDeepLearning2022,laviSurveyReliableDeep2020,mingDeepLearningbasedPerson2022,pengDeepLearningBasedOccluded2023} summarized deep learning based pedestrian re-recognition methods. \cite{wangCrossDomainPersonReidentification2021, singhComprehensiveSurveyPerson2022} investigated the available image/video-based person Re-ID datasets. \cite{zhengPersonReidentificationPresent2016,zahraPersonReidentificationRetrospective2023} surveyed image-based or video-based pedestrian re-recognition systems. \cite{karanamSystematicEvaluationBenchmark2019,mingDeepLearningbasedPerson2022,zouPersonReidentificationBased2021,yeDeepLearningPerson2022} proposed metric learning based classification method for pedestrian re-recognition. 
\cite{islamPersonSearchNew2020,yeDeepLearningPerson2022,mingDeepLearningbasedPerson2022} presented a pedestrian re-recognition taxonomy based on representation learning. Huang et al.\cite{huangDeepLearningVisibleinfrared2023} provided a visible-infrared cross-modal categorization of pedestrian re-identification studies. Wang et al.\cite{wangIntramodalitySurveyHeterogeneous2020} reviewed cross-modality based person Re-ID research from four aspects, including sketch, text, low resolution (LR) and infrared (IR), as Figure \ref{fig2} is introduced later.

However, these surveys exhibit areas for improvement, such as lacking a systematic categorization and analysis of text-based deep learning methods in person re-identification, and omitting a comprehensive discussion of text-based person re-identification. Therefore, in comparison to the aforementioned reviews, this paper delves into an in-depth analysis of recent research in text-based person re-identification from dimensions including datasets, strategies, architectures, and optimizations. We comprehensively review existing deep learning-based approaches and discuss their strengths and limitations.


\section{Evaluation}
The evaluation of text-based person Re-ID tasks, typically involves publicly available benchmark datasets and common performance metrics. This provides a standardized testing platform, enabling a quantified and objective comparative assessment of the effectiveness of different experimental algorithms.

\subsection{Datasets}

\begin{table*}
  \centering
  \caption{The table summarizes the available public datasets for attribute-based and NL-based person Re-ID. The datasets are summarized in terms of dimensions such as dataset, number of pedestrians, attributes/description, and data source, respectively.}
  \label{dataset}
  \begin{tabular}{c|c|c|c|c|c}
  \hline
   & \textbf{Dataset} & \textbf{Pedestrians} & \multicolumn{2}{c|}{\textbf{Attribute/Description}}  & \textbf{Source} \\
   \multirow{8}{*}{Attribute-based}& & & Binary & Multi-class & Outdoor/Indoor \\
  \hline
  \hline
  &PETA[9] & 19000 & 61 & 4 & Indoor \\
  &RAP[88] & 41585 & 69 & 3 & Indoor \\
  &RAP V2.0[89] & 84928 & 69 & 3 & Outdoor \\
  &PA-100K[90] & 100000 & 26 & / & Outdoor \\
  &Market1501-Attribute[8] & 32668 & 26 & 1 & Outdoor \\
  &DukeMTMC-Attribute[8] & 34183 & 23 & / & Outdoor \\
  &UAV-Human[87] & 22263 & 7 & / & Outdoor \\
  \hline
  \multirow{4}{*}{Natural Language-based} & & & \multicolumn{2}{c|}{Sentences * words} & \\
  &CUHK-PEDES[4] & 34054 & 2 & $>23$ & Outdoor+Indoor \\
  &RSTPReid[24] & 20505 & 2 & $>25.8$ & Outdoor+Indoor \\
  &ICFG-PEDES[65] & 54522 & 3 & $>37.2$ & Outdoor+Indoor \\
  \hline
  \end{tabular}
\end{table*}



\subsubsection{attribute-based Person Re-ID Datasets}
Attribute-based Person Re-identification (Re-ID) often involves the utilization of semantic attributes that describe appearance characteristics of individuals, such as gender, age, clothing type, clothing texture, and clothing color \cite{vaqueroAttributebasedPeopleSearch2009}. However, these attributes frequently encompass low-level semantic attributes that can be directly associated with image regions, such as clothing type, clothing texture, and clothing color, as well as high-level semantic attributes that directly correspond to people, such as gender and age. By leveraging these attributes, the negative impact of variations in viewpoint and appearance on person re-identification can be mitigated, resulting in more reliable recognition and tracking performance \cite{jeongASMRLearningAttributeBased2021}. To facilitate research in Attribute-based Person Re-ID, researchers have initiated the construction of several datasets specifically tailored for this purpose. Notable publicly available datasets in this domain include PETA \cite{dengPedestrianAttributeRecognition2014a}, UAV-Human \cite{liUAVHumanLargeBenchmark2021}, RAP ~\cite{liRichlyAnnotatedDataset2016a}, RAP2.0 \cite{liRichlyAnnotatedPedestrian2019}, PA-100K \cite{liuHydraPlusNetAttentiveDeep2017}, Market1501-Attribute \cite{linImprovingPersonReidentification2019}, and DukeMTMC-Attribute \cite{linImprovingPersonReidentification2019}. 

These datasets typically consist of a substantial collection of images capturing individuals from surveillance cameras, each accompanied by attribute labels. The dataset creation process generally involves selecting a subset of images from existing Person Re-ID datasets. Subsequently, attribute labels are assigned to each image either manually or through automated annotation. Ultimately, the annotated images along with their corresponding attribute labels are organized into a unified dataset, which serves as a resource for researchers in the field, as shown in Table~\Ref{dataset}.

\subsubsection{Natural Language-based Person Re-ID Datasets}
Publicly available NL-based person re-identification datasets have significant practical implications for benchmarking, algorithm improvement, security and surveillance applications, privacy protection, and advancing multimodal research. Currently, there are three large publicly available datasets, CUHK-PEDES, RSTPReid, and ICFG-PEDES. The details of these datasets will be described. 

\paragraph{CUHK-PEDES}
CUHK-PEDES is the first description dataset introduced by The Chinese University of Hong Kong \cite{liPersonSearchNatural2017}, designed for training and evaluating person description and image retrieval tasks in computer vision. The dataset was collected by aggregating person images from five existing and well-known pedestrian re-identification evaluation datasets, namely CUHK03 \cite{liDeepReIDDeepFilter2014}, Market-1501 \cite{zhengPersonReidentificationMeets2015}, SSM \cite{xiaoJointDetectionIdentification2016b}, VIPER \cite{grayEvaluatingAppearanceModels2007}, CUHK01 \cite{liHumanReidentificationTransferred2013}, and subsequently annotated with descriptive text by crowdsourced workers from Amazon Mechanical Turk (AMT). It consists of over 40,206 images with 80,412 person descriptions, each containing at least 23 words, across 13,003 unique person identities. The training set includes 34,054 images, 11,003 unique persons, and 68,108 textual descriptions. The validation set contains 3,078 images, 1,000 unique persons, and 6,158 textual descriptions, while the test set comprises 3,074 images, 1,000 unique persons, and 6,156 textual descriptions. The individuals in the dataset were captured in street and public places in the Hong Kong area. CUHK-PEDES provides diverse person descriptions encompassing appearance, actions, postures, and interactions, with each person description linked to the corresponding image ID. The release of the CUHK-PEDES dataset promotes interdisciplinary research between computer vision and natural language processing, providing researchers in the fields of person description and image retrieval with the first benchmark dataset.

\paragraph{RSTPReid}
The RSTPReid dataset is a pedestrian re-identification dataset introduced by Nanjing University of Science and Technology \cite{zhuDSSLDeepSurroundingsperson2021}, designed for training and evaluating pedestrian re-identification tasks in computer vision. To address the issue in the CUHK-PEDES dataset where each specific pedestrian is captured by the same camera under the same time-space conditions, which does not reflect real-world scenarios, the authors constructed the Real Scenarios Text-based Person Reidentification (RSTPReid) dataset based on MSMT17 \cite{weiPersonTransferGAN2018a}. The RSTPReid dataset contains 20,505 images of 4,101 individuals captured by 15 independent cameras from different viewpoints, lighting conditions, locations, and weather conditions. Each individual has 5 corresponding images taken by different cameras, and each image is accompanied by 2 text descriptions with no fewer than 23 words. The training set includes 3,701 identities, the validation set has 200 identities, and the test set also consists of 200 identities. The dataset includes a relatively small number of pedestrian identities but covers both indoor and outdoor scenarios. Each pedestrian appears in multiple images with different shooting angles and lighting conditions, making this dataset more challenging and realistic.

\paragraph{ICFG-PEDES}
The Identity-Centric and Fine-Grained Person Description Dataset (ICFG-PEDES) is a person description dataset introduced by South China University of Technology \cite{dingSemanticallySelfAlignedNetwork2021}, which also serves as a new benchmark dataset for research in person description and image retrieval domains. Similar to CUHK-PEDES, this dataset contains a large number of person descriptions paired with corresponding image IDs. ICFG-PEDES consists of 54,522 pedestrian images of 4,102 distinct identities, all collected from the MSMT17 database \cite{weiPersonTransferGAN2018a}. The training set comprises 34,674 image-text pairs of 3,102 pedestrians, while the test set contains 19,848 image-text pairs of 1,000 pedestrians. Each image is associated with only one text description, with an average of 37.2 words per description. Compared to the CUHK-PEDES dataset, ICFG-PEDES emphasizes more on fine-grained person descriptions and reduces some irrelevant action and background information. Additionally, it addresses the issue of consistent backgrounds in the former dataset by emphasizing more on appearance variations. Consequently, the ICFG-PEDES dataset can be employed for more challenging image retrieval and person description tasks.

\subsection{Metrics}

\subsubsection{Mean Average Precision(mAP)}

For each query i, we define a precision $P_i(j)$ as the proportion of correct matches in the top j matches, and then for each positive instance, we calculate the average of the proportions of all positive instances before this one, which is the Average Precision $AP_i$:

\begin{equation}
\label{eq1}
AP_i = \frac{1}{M_i} \sum_{j=1}^{M_i} P_i(j) \cdot I(rank_i = j).
\end{equation}

Where $M_i$ is the number of positive instances for query i, and $\sum$ is the summation over all positive instances. The symbol \( I(\text{rank}_i = j) \) usually denotes an indicator function. The indicator function takes the value of 1 under certain conditions, otherwise it takes the value of 0. Specifically:
\[ I(\text{rank}_i = j) \] indicates that the indicator function \( I(\text{rank}_i = j) \) takes the value of 1 when the first \(i\) query's first \(j\) result is a correct match, otherwise it takes the value of 0. In other words, it determines whether or not there is a correct match in the ranked position of \(j\).
Then, mAP is the average of the Average Precision of all queries:

\begin{equation}
\label{eq2}
mAP = \frac{1}{N} \sum_{i=1}^{N} AP_i.
\end{equation}

Where $N$ is the total number of queries, and $\sum$ is the summation over all queries.

\subsubsection{Rank-N Accuracy}
For each query i, we define an indicator function $I(rank_i \leq n)$, which is 1 if the correct match is within the top n matches, and 0 otherwise. Then, Rank-n accuracy is the average of this function:

\begin{equation}
\label{eq3}
Rank-n = \frac{1}{N} \sum_{i=1}^{N} I(rank_i \leq n).
\end{equation}

Where $N$ is the total number of queries, and $\sum$ is the summation over all queries.


\section{Strategy}
In the task of text-based person re-identification, image data is typically captured through CCTV in open environments [45]. This data not only contains key descriptive features of individuals but also includes a significant amount of background noise. Early works \cite{liPersonSearchNatural2017,zhangDeepCrossModalProjection2018a, yanCrossdomainPersonReidentification2022b,chenImprovingTextBasedPerson2018} typically directly extract global image text features for simple cross-modal matching. However, due to the uniqueness of the person re-identification task, which differs from existing image-text retrieval \cite{zhengDualpathConvolutionalImageText2020,jingPoseGuidedMultiGranularityAttention2020a}, the categories to be retrieved all belong to highly similar individuals. Identification based solely on distinguishing features such as clothing and posture increases the challenges of the text-based person re-identification task.

\subsection{Stripe Segmentation}
Some work suggests that various parts of the human body are evenly arranged in images, and Stripe Segmentation can be employed as a guiding model to achieve multimodal local alignment supervision \cite{niuImprovingDescriptionBasedPerson2020,jiAsymmetricCrossScaleAlignment2022}. Stripe Segmentation is a commonly used technique in pedestrian re-identification, aiming to mitigate the impact of pose, occlusion, and lighting variations on re-identification performance within pedestrian images \cite{niuImprovingDescriptionBasedPerson2020}, \cite{dingSemanticallySelfAlignedNetwork2021}. The fundamental idea is to partition pedestrian images into multiple vertical stripes, also referred to as person parts. Each stripe can be considered a subregion of the pedestrian image (head, upper body, lower body, foot). The advantage of this approach lies in dividing pedestrian images into multiple local regions, effectively capturing local features of pedestrians to enhance recognition accuracy \cite{liTransformerBasedLanguagePersonSearch2021}. For instance, local texture, color, and shape features can be extracted within each stripe, facilitating better differentiation between different parts of pedestrians. Stripes of different heights can correspond to different scales of pedestrian parts, providing richer scale information to address scale variation issues in pedestrian images. Ding et al. \cite{dingSemanticallySelfAlignedNetwork2021} proposed a Semantic Self-Alignment Network (SSAN) capable of automatically extracting part-level textual features for corresponding visual regions. They designed a multi-view non-local network to capture relationships between body parts, thereby establishing better correspondences between body parts and noun phrases. Li et al \cite{liTransformerBasedLanguagePersonSearch2021} proposed to vertically divide a character image into multiple regions using two different approaches, including overlapping slices and key-point-based slices. Niu et al. \cite{niuImprovingDescriptionBasedPerson2020} also emphasized local-local alignment, employing Stripe Segmentation in the proposed Bidirectional Fine-Grained Matching (BFM) module to match visual human body parts with noun phrases. However, this strategy is fragile and sensitive to conditional changes. For example, the regions of image samples do not always contain a complete human body, and at times, the human head does not consistently appear in the first stripe, significantly impacting the robustness of existing methods.

\subsection{Multi-scale Fusion}
Due to the small inter-class variance in both pedestrian images and descriptions, comprehensive information is required for pedestrian retrieval to coordinate visual and textual clues across all scales. Subsequent to the introduction of local feature extraction strategies other than stripe segmentation, the utilization of fused features is referred to as multi-scale feature fusion \cite{wangPartBasedMultiScaleAttention2022a}, as illustrated in Figure ~\ref{fig4}. Image Structure Graph Network (A-GANet) \cite{liuDeepAdversarialGraph2019} employs residual modules for visual representation extraction and utilizes designed graph attention convolutional layers for structured high-level visual semantic feature extraction. Specifically, it captures relationship features of different pedestrian parts through graph convolutional networks. Wang et al. \cite{wangViTAAVisualTextualAttributes2020a} proposed the Visual Text Attribute Alignment model (ViTAA), which learns to decompose the person's feature space into subspaces corresponding to attributes using an optically assisted attribute segmentation layer. On the image side, a segmentation layer is employed to divide pedestrian images into full body, head, upper clothes, Lower clothes, shoes, and bag parts. On the text side, corresponding phrases of different segmented parts are extracted. 

Ji et al. \cite{jiAsymmetricCrossScaleAlignment2022} introduced an Asymmetric Cross-Scale Alignment (ACSA) method. Global text representation and local phrase representation are employed for text extraction. The global representation is divided into four regions as local visual representations, namely Head, Upper body, Lower body, and foot. These features are used afterwards for feature concatenation and computation. The partition strategy does not involve additional computational costs but can better retain prominent body parts for fine-grained matching. Wang et al.\cite{wangTextbasedPersonSearch2021} proposed a novel Multi-Granularity Embedding Learning (MGEL) model. It generates multi-granularity embeddings of partial human bodies from coarse to fine by revisiting person images at different spatial scales. Chen et al.\cite{chenTIPCBSimpleEffective2022} adopted a multi-branch representation in the learning path, enabling text-adaptive matching of corresponding visual local representations, thereby aligning text and visual local representations. Additionally, a multi-stage cross-modal matching strategy was proposed to eliminate modality gaps from low-level, local, and global features, gradually narrowing the feature gap between image and text domains. 

Niu et al.\cite{niuImprovingDescriptionBasedPerson2020} presented a Multi-Granularity Image-Text Alignment (MIA) framework to address cross-modal fine-grained issues. It consists of three modules corresponding to three granularities. The Global Contrast (GC) module is used for global-global alignment. The Relation-Guided Global-Local Alignment (RGA) module filters global-local relationships. The Bidirectional Fine-Grained Matching (BFM) module aligns local to local based on trained fine-grained local components. Zheng et al. \cite{zhengHierarchicalGumbelAttention2020} proposed a Hierarchical Adaptive Matching model to learn subtle feature representations from three different granularities (i.e., word-level, phrase-level, and sentence-level) for fine-grained image-text retrieval tasks. Wang et al. \cite{wangDivideandMergeEmbeddingSpace2021} addressed the issue that existing methods typically learn similar mappings between local parts of images and text or embed entire images and text into a unified embedding space, resulting in a problem of local-global correlation. They designed a Divide and Merge Embedding (DME) learning framework for text-based person search. It models the relationship between local parts and global embedding, merging local details into the global embedding. Li et al.\cite{liPersonTextImageMatching2022} constructed features for text and person image matching, different from other methods. The latter acquires global features by aggregating aligned local features, leading to misalignment between text and visual features. They proposed placing features with the same semantics in the same spatial position to construct the final features for person-text-image matching, achieving semantic consistency and interpretability of global features. 

Yan et al.\cite{yanImageSpecificInformationSuppression2022} proposed an Implicit Local Alignment module, adaptively aggregating image and text features into a set of modality-shared semantic topic centers, implicitly learning local fine-grained correspondence between images and text without additional supervision information and complex cross-modal interaction. Additionally, global alignment is introduced as a supplement to the local perspective. Gao et al. \cite{gaoContextualNonLocalAlignment2021} introduced a method called Non-Local Alignment on Full Scale (NAFS), capable of adaptively aligning image and text features at all scales. First, a novel ladder network structure is proposed to extract full-size image features with better locality. Secondly, a language model, Bidirectional Encoder Representations from Transformers (BERT), with local constraint attention is proposed to obtain description representations at different scales. Then, instead of aligning features separately at each scale, a novel context non-local attention mechanism is applied to simultaneously discover potential alignment at all scales. 

In summary, applying target detection or additional branch networks to detect significant regions and then extract local features for obtaining multi-scale features has certain accuracy advantages. However, due to the increased external network overhead, these methods often result in higher computational costs.

\subsection{Attention Mechanism}
Cross-modal alignment is inherently challenging in achieving fine-grained matching between text and images. In addition to implicit alignment strategies achieved through the automatic fusion of multi-scale features, another key approach involves the use of explicit alignment strategies. For instance, employing attention mechanisms enhances the capability of feature extraction by capturing discriminative feature representations related to both language descriptions and visual appearances. This aids in intuitively understanding and controlling the alignment process. Attention models can establish correspondences between body parts and words \cite{liPersonSearchNatural2017,liIdentityAwareTextualVisualMatching2017,yanCrossdomainPersonReidentification2022b,wangTextbasedPersonSearch2021,dingSemanticallySelfAlignedNetwork2021,farooqAXMNetImplicitCrossModal2022,zhangTextbasedPersonSearch2021}, offering the advantage of not relying on external cues for aligning feature embeddings. These attention strategies can be categorized into spatial attention, channel attention, mixed attention, non-local attention, and positional attention.

Liu et al. \cite{wangTextbasedPersonSearch2021} proposed an attention-based deep neural network capable of capturing multiple attention features from low-level to semantic layers to learn comprehensive features for fine-grained pedestrian representation. Wang et al. \cite{liuEfficientTextbasedPerson2022} utilized a multi-head self-attention module to extract embeddings for different granularity portions of the text stream, employing adaptive filtering to remove interference and obtain fine-grained features. Lee et al. \cite{leeStackedCrossAttention2018} introduced a Stacked Cross Attention Mechanism (SCAN) that compares attention information for text and image in two stages, determining the importance of image regions and words to achieve potential alignment between the two modalities. Spatial attention-based methods often focus solely on local discriminative features of pedestrians, neglecting the impact of feature diversity on pedestrian retrieval.

Li et al. \cite{liHybridAttentionNetwork2020} proposed a cubic attention convolutional neural network that combines spatial and channel attention to maximize complementary information from different scale attention features, addressing the challenges of cross-modal alignment in pedestrian re-identification. Wang et al. \cite{wangDivideandMergeEmbeddingSpace2021} designed a Feature Division Network (FDN) that embeds inputs into K locally guided semantic representations using self-attention, each representing different parts of a person. They then introduced a Correlation-based Subspace Projection (RSP) method to merge different local representations into a compact global embedding. Yan et al. \cite{yanImageSpecificInformationSuppression2022} presented an Efficient Joint Information and Semantic Alignment Network (ISANet) for text-based person search. Specifically, they designed an image-specific information suppression module that suppresses image background and environmental factors through relationship-guided localization and channel attention filtering, effectively alleviating information inequality and achieving information alignment between text and image.

Considering the indistinguishability of features extracted from first-order attention (e.g., spatial and channel attention) in complex camera views and pose-changing scenarios, Farooq et al. \cite{farooqAXMNetImplicitCrossModal2022} applied non-local attention directly after calculating interactions between text features to model long-term dependencies between different phrases in the text. Inspired by self-attention, Gao et al. \cite{gaoContextualNonLocalAlignment2021} proposed a Context Non-Local Attention to enable cross-modal features to align with each other in a coarse-to-fine manner based on their semantics, rather than relying solely on predefined and fixed rules (e.g., local-local, global-global). Wang et al. \cite{wangPartBasedMultiScaleAttention2022a} introduced a Part-based Multi-scale Attention Network (PMAN), consisting of a dual-path feature extraction framework with attention-based branches for a multi-scale cross-modal matching module. This module extracts visual semantic features from different scales and matches them with text features.

Addressing concerns that utilizing soft attention mechanisms to infer semantic alignment between image regions and corresponding words in a sentence may lead to the fusion of unrelated multimodal features and result in redundant matching, Wang et al. \cite{wangIMGNetInnercrossmodalAttentional2020e} proposed an IMG-Net model. This model combines intra-modal self-attention and cross-modal hard region attention with a fine-grained model to extract multi-granularity semantic information. Zheng et al. \cite{zhengHierarchicalGumbelAttention2020} introduced a Hierarchical Gumbel Hard Attention Module, using the Gumbel top-k reparameterization algorithm as a low-variance, unbiased gradient estimator to select strongly semantically related regions for all regions of the image and corresponding words or phrases, enhancing precise alignment and similarity calculation results.

While these techniques often bring about noticeably superior retrieval performance, they frequently come at the cost of efficiency. Specifically, for $M$ galleries and $N$ queries, their complexity is $O(M+N)$. In contrast, the complexity of attention-based cross-modal approaches rises to $O(MN)$ due to pairwise inputs. To address this issue, Yang et al. \cite{yangUnifiedTextbasedPerson2023a} initially employ cross-modal non-attention features to quickly identify candidates and then deploy attention-based modules to refine the final ranking scores.

\subsection{External Auxiliary}
For Text-based Person Re-Identification (Re-ID) tasks, several auxiliary strategies such as image segmentation, random mask, human body keypoints, attribute prediction, clustering analysis, and color extraction can be employed to extract advanced semantics, thereby improving cross-modal retrieval performance. \cite{reedLearningDeepRepresentations2016} proposed a dual-part simultaneous alignment representation scheme using attention mechanisms to capture distinguishing information beyond body parts by leveraging complementary information from accurate human parts and noisy parts to update representations.

To leverage multi-level visual content, Jing et al. \cite{jingPoseGuidedMultiGranularityAttention2020a} introduced a Pose-guided Multi-granularity Attention Network (PMA). They initially proposed a Coarse Alignment network (CA) that uses similarity-based attention to select relevant image regions for global description. Pose information can be utilized to learn potential semantic alignments between visual body parts and textual noun phrases. Shu et al. \cite{shuSeeFinerSee2023} introduced Bidirectional Mask Modeling (BMM) without the need for additional manual labeling. They applied random masks to images and corresponding text keywords, forcing the model to explore more useful matching clues. This approach increases data diversity and enhances model generalization. 

Wang et al. \cite{wangViTAAVisualTextualAttributes2020a} proposed a novel Visual-Textual Attribute Alignment model (ViTAA). For semantic feature extraction, segmentation labels are used to drive the learning of attribute-aware features from input images. Aggarwal et al. \cite{aggarwalTextbasedPersonSearch2020a} introduced a method for text-based person search by learning attribute-driven spatial and class-driven spatial information. The goal is to create semantically retained embeddings through an additional task of attribute prediction. Suo et al. \cite{suoSimpleRobustCorrelation2022a} proposed a novel Simple Robust Correlated Filtering (SRCF) framework, which effectively extracts key clues and adaptively aligns multimodal features by constructing a set of universal semantic templates (filters). This framework differs from previous attention-based methods, focusing on computing similarities between templates and inputs. Two types of filtering modules (denoising filter and dictionary filter) were designed to extract key features and establish multimodal mappings.

Zhao et al. \cite{zhaoWeaklySupervisedTextBased2021} introduced a Cross-Modal Mutual Training (CMMT) framework. Specifically, to alleviate intra-class variation, clustering methods generate pseudo-labels for visual and text instances. To further refine clustering results, CMMT provides a mutual pseudo-label refinement module, which utilizes clustering results in one modality to refine clustering results in another modality constrained by text-image pairwise relationships. Chen et al. \cite{chenIntegratingInformationTheory2021} integrated information theory and adversarial learning into an end-to-end framework, exploring information theory to reduce heterogeneity gaps in cross-modal retrieval. This method is beneficial for building a shared space to further learn commonalities between cross-modal features, applicable to video-text matching. Additionally, regularization terms based on KL-divergence and temperature scaling were introduced to address the issue of data imbalance. 

Li et al. \cite{liJointTokenFeature2022a} proposed a novel Joint Label and Feature Alignment Framework (TFAF) to gradually reduce inter-modal and intra-class gaps. Firstly, a dual-path feature learning network is constructed to extract features and align them to reduce inter-modal gaps. 
Secondly, a text generation module is designed to generate label sequences using visual features, followed by label alignment to reduce intra-class gaps. Finally, a fusion interaction module is introduced, utilizing a multi-stage feature fusion strategy to further eliminate modality heterogeneity. Chen et al. \cite{chenCrossModalKnowledgeAdaptation2021} introduced the concept of knowledge distillation to image and text networks, aiming to improve the matching capabilities of two networks with different modalities by balancing information from both networks. 
Wang et al. \cite{wangCAIBCCapturingAllround2022} addressed the color-dependency issue in cross-modal pedestrian retrieval by proposing a Color-Aware Information Beyond Color(CAIBC) architecture through joint optimization of a multi-branch structure. 
CAIBC includes three branches(RGB, Grayscale, and Color) employing mutual learning mechanisms for knowledge exchange across branches. 
It achieved an accuracy of 88.37\% in top-10 on the CUHK-PEDES dataset for weakly supervised text-to-person retrieval.


\section{Architecture}
When mapping features from a specific modality to a common manifold, the feature distribution of other modalities remains imperceptible. This implies that embedding and aligning multimodal features in the common manifold entirely depend on the model's own experience rather than the actual data distribution. In other words, a major challenge in cross-modal Person Re-ID is ensuring that the feature distribution in the common manifold accurately reflects the feature distribution in the original modalities. This necessitates the model to possess sufficient capability to capture and understand the relationships between different modalities, which typically requires abundant data and carefully designed model structures.

\subsection{Convolutional Neural Network}
Text-based person Re-ID can be viewed as a multimodal pedestrian retrieval problem, with the primary challenge lying in the extraction of features from both visual and textual data. Due to the capability of neural networks to automatically learn and extract features from pedestrian images, the cumbersome process of manual feature design can be avoided. Most existing efforts consider employing network architectures designed for image classification as the backbone of visual networks. Consequently, convolutional neural networks (CNNs) have been utilized extensively in text-based pedestrian re-identification, primarily leveraging CNNs \cite{chenTIPCBSimpleEffective2022,niuCrossmodalCooccurrenceAttributes2022,farooqAXMNetImplicitCrossModal2022,liuEfficientTextbasedPerson2022,farooqConvolutionalBaselinePerson2020,CascadeAttentionNetwork2018,farooqCrossModalPerson2020,geVisualTextualAssociationHardest2019a}, and their major variants such as VGG-16 \cite{reedLearningDeepRepresentations2016, vinyalsShowTellNeural2015, liPersonSearchNatural2017,liIdentityAwareTextualVisualMatching2017,chenImprovingTextBasedPerson2018, niuImprovingDescriptionBasedPerson2020, wangImprovingEmbeddingLearning2022}, MobileNet \cite{wangLanguagePersonSearch2019, aggarwalTextbasedPersonSearch2020a, wuLapsCoreLanguageGuidedPerson2021, zhangDeepCrossModalProjection2018a}, ResNet50 \cite{wangAMENAdversarialMultispace2021,suoSimpleRobustCorrelation2022a,liHybridAttentionNetwork2020,wangCAIBCCapturingAllround2022,farooqCrossModalPerson2020}, ResNet101 \cite{sarafianosAdversarialRepresentationLearning2019, niuFusingTwoDirections2019a,niuTextualDependencyEmbedding2020,liTransformerBasedLanguagePersonSearch2021, hanTextBasedPersonSearch2021a}, ResNet-152 \cite{chenIntegratingInformationTheory2021}, etc., for extracting crucial information in visual features. These networks are sometimes also employed for text feature extraction \cite{farooqAXMNetImplicitCrossModal2022,farooqConvolutionalBaselinePerson2020,farooqCrossModalPerson2020,niuCrossmodalCooccurrenceAttributes2022}. The combination of pedestrian images extracted by CNNs and textual description information is achieved, aligning low and high-level semantic information across modalities in an implicit space using matching methods. This facilitates the recognition and matching of pedestrian identities.

Two independently pre-trained ResNet-50 models on ImageNet serve as the visual backbone in study \cite{wangCAIBCCapturingAllround2022}. Addressing the challenge of manual annotation for existing cross-modal data, Jing et al. propose a Matrix Alignment Network (MAN) to tackle cross-modal cross-domain person search tasks, applying the single-modal cross-domain adaptive idea to the problem of image-text-based cross-modal person re-identification \cite{jingCrossModalCrossDomainMoment2020}. Farooq et al. introduce the Deep Vision Net model based on ResNet-50 to construct the visual feature extraction backbone \cite{farooqCrossModalPerson2020}. Deviating from the original architecture, the proposed network includes two fully connected layers before the classifier layers, applies batch normalization after the two FC layers, and employs dropout before the final layer. They further present a two-stream deep convolutional neural network framework supervised by cross-entropy loss \cite{farooqConvolutionalBaselinePerson2020}. The language network is modified to resemble the visual branch in terms of the number of layers in deep residual networks, connecting the class probability weights between the second-to-last layer and the final layer, i.e., sharing logits of the softmax layer across both networks. AXM-Net, proposed by \cite{farooqAXMNetImplicitCrossModal2022}, is a novel architecture based on convolutional neural networks (CNN) that does not rely on external cues for explicit alignment of feature embeddings and is capable of learning semantically aligned cross-modal feature representations. Most of the aforementioned works initialize pre-trained ResNet-50 \cite{heDeepResidualLearning2016} and ResNet-101 \cite{niuFusingTwoDirections2019a} backbones on large datasets to better map modal-corresponding information. Additionally, the application of large models such as CLIP in some fine-grained visual tasks has demonstrated effective paradigm transfer, fully migrating visual-text large models based on contrastive learning to text-based person re-identification tasks \cite{yanCLIPDrivenFinegrainedTextImage2022}. Presently, many studies commonly regard representation learning as a classification method for person re-identification, believing that feature embedding is a core module for text-based person re-identification tasks, crucial for narrowing the semantic gap between modalities. CNNs exhibit significant advantages in perceptual localization of local features in pedestrian vision and end-to-end learning. However, they still have certain limitations in handling long-term dependent data relationships.

\subsection{Recurrent Neural Network}
Beyond visual features, textual descriptions directly impact the accuracy of text-based Person Re-ID. Efficient extraction of textual features is crucial under existing textual data descriptions \cite{dingSemanticallySelfAlignedNetwork2021}. Recurrent Neural Networks (RNNs) and their variants, as neural networks for processing sequential events, including text, have been widely accepted in the research community \cite{vinyalsShowTellNeural2015,reedLearningDeepRepresentations2016,wangImprovingEmbeddingLearning2022,liPersonSearchNatural2017}. Early on, in this field, word embedding techniques were initially employed to map words to vector spaces. Subsequently, variants such as LSTM \cite{liIdentityAwareTextualVisualMatching2017}, Bi-LSTM \cite{wangViTAAVisualTextualAttributes2020a, chenImprovingDeepVisual2018,sarafianosAdversarialRepresentationLearning2019,zhangDeepCrossModalProjection2018a,niuImprovingDescriptionBasedPerson2020, wangLanguagePersonSearch2019, jingPoseGuidedMultiGranularityAttention2020a, chenCrossModalKnowledgeAdaptation2021, chenIntegratingInformationTheory2021,zhaoWeaklySupervisedTextBased2021, wangTextbasedPersonSearch2021, yanImageSpecificInformationSuppression2022} attempted to capture long dependencies of sentences for semantic feature extraction from text sequences. Later, with the emergence of large language models like Bidirectional Encoder Representations from Transformers (BERT)\cite{devlinBERTPretrainingDeep2019}, the paradigm shift of "pre-training + fine-tuning" has been triggered in the Text-based Person Re-ID domain. Utilizing pre-trained models to learn general features of natural language and obtaining high-quality word embeddings, followed by inputting into RNNs variants to capture contextual dependencies.

Niu et al. \cite{niuFusingTwoDirections2019a} first proposed a concise and effective framework for image-text alignment to address visual-semantic differences. Innovatively integrating two opposing directions, source to target and target to source, for cross-domain adaptation. Wang et al. \cite{wangAMENAdversarialMultispace2021} introduced a novel Adversarial Multi-space Embedding Network (AMEN) to learn and match embeddings in multiple spaces. Bi-GRU is used for text feature extraction, and the paradigms of inter-modal and intra-modal reconstruction collaborate to embed features correctly into the opposite modal space while learning a robust common space. \cite{liuDeepAdversarialGraph2019} introduces a Text-Graph Attention Network consisting of a Bi-LSTM layer, a text scene graph module, and a joint embedding layer. Bi-LSTM units gradually take the embeddings of each word in the text as input, capturing the historical and future context information of processed words. Yan et al. \cite{yanImageSpecificInformationSuppression2022} employed an LSTM-based Recurrent Feature Aggregation Network to accumulate discriminative features from the first LSTM node to the deepest LSTM node, effectively alleviating interference caused by occlusion, background noise, and detection failures. Chen et al. \cite{chenImprovingTextBasedPerson2018} decomposed video sequences into multiple segments and used LSTM to learn to detect the segments where the images are located in both time and space features. This approach reduces variations in the same person across samples, favoring the learning of similarity features. Both of the above methods independently process each video frame. Ding et al. \cite{dingSemanticallySelfAlignedNetwork2021} employed Bidirectional Long Short-Term Memory Network (Bi-LSTM) to process each text description, capturing relationships between words. With contextual cues, the Word Attention Module (WAM) based on word representations infers word part correspondences, referring to word part correspondences to obtain original part-level texture features. Wang et al. \cite{wangLookYouLeap2022} proposed a new algorithm called LBUL to learn a Consistent Cross-Modal Common Manifold (C3M) for text-based person retrieval. For textual samples, the proposed method embeds the text modality into C3M after considering the distribution features of cross-modalities comprehensively to achieve more reliable cross-modal distribution consensus, rather than blind predictions. However, RNN and its variants also have some issues in the Text-based Person Re-ID task. For instance, the embedding quality of pedestrian text inputs is often very sensitive, features extracted are typically influenced by the length of data sequences, and RNNs only establish temporal correlations on high-level features, thus failing to capture the temporal clues describing the local details of images.

\subsection{Autoencoder}
Sometimes, text-based Person Re-ID tasks also involve the utilization of Autoencoder network architectures. Commonly, these models map images or textual descriptions into a low-dimensional embedding space and employ this embedding space for person re-identification tasks \cite{liIdentityAwareTextualVisualMatching2017,hanTextBasedPersonSearch2021a}. In this manner, pedestrian images and their corresponding textual descriptions are mapped to a low-dimensional vector space, which encapsulates crucial semantic information, as both image and text provide pivotal details about the identity of the pedestrian. Autoencoders, particularly those with atypical encoder components, exhibit several advantages in Text-based Person Re-ID tasks. Firstly, they can overcome issues prevalent in traditional approaches arising from the lack of semantic information in image-text pairs. Secondly, Autoencoder models proficiently capture essential information from textual descriptions and compress it, facilitating effective data compression. Additionally, this approach boasts good interpretability, as relationships between pedestrian images and textual descriptions can be visualized based on vectors within the embedding space. Wang et al. \cite{wangAMENAdversarialMultispace2021} introduced an innovative Adversarial Multi-space Embedding Network (AMEN) for learning and matching embeddings across multiple spaces. Following an encoder-decoder paradigm, the inter-modal reconstruction paradigm collaborates with the intra-modal reconstruction paradigm to embed features accurately into the opposite modal space while learning a robust common space. \cite{wangImprovingEmbeddingLearning2022} proposed an Improved Embedding Learning through Virtual Attribute Decoupling (iVAD) model. This method enhances embedding learning by using an encoder-decoder structure to decompose attribute information in images and text. Pedestrian attributes are treated as hidden vectors, and attribute-related embeddings are obtained. Diverging from previous works that separate attribute learning from image-text embedding learning, a hierarchical feature embedding framework is introduced. The Attribute-Enhanced Feature Embedding (AEFE) module merges attribute-related embeddings into the learned image-text embeddings, leveraging attribute information to enhance feature discriminability. However, a drawback of this approach is its reliance on a substantial amount of training data to train the Autoencoder model, and meticulous tuning of model parameters and embedding space dimensions is necessary to achieve optimal performance.

\subsection{Graph Neural Network}
Utilizing Graph Neural Networks (GNN) for text-based pedestrian re-identification has become a noteworthy research direction in recent years \cite{liuDeepAdversarialGraph2019,liPersonTextImageMatching2022}. GNN is a type of deep learning method based on graph structures, suitable for handling unstructured data and relationships, and applicable to various data types such as images, text, and structured data. In text-based Person Re-ID tasks, GNN is often employed to construct a graph structure, where each node represents a pedestrian body part along with its corresponding phrase description. Edges between nodes represent semantic relationships between different body parts and descriptions, such as verb (wear), noun (women), adjective (yellow), and others. The methods utilizing GNN for text-based Person Re-ID typically involve the following steps. First, images and corresponding text descriptions for each pedestrian are extracted from the dataset. Second, two graph convolution structures are established based on auxiliary strategies, such as attention mechanisms, image segmentation, etc. Afterwards, Graph Convolutional Networks (GCN) are employed to learn the relationships between body nodes in the graph structure and transform these relationships into feature vectors. Finally, feature vectors of body structures undergo cross-modal asymmetric alignment, and the text modality is input into a classifier for Person Re-ID.

The methods employing GNN for text-based Person Re-ID offer several advantages. Firstly, they can capture semantic relationships between different pedestrian images and text descriptions, facilitating future inference-based retrieval, i.e., recognizing a pedestrian engaged in complex events. Secondly, GNN can handle text descriptions of varying lengths and structures, providing the advantage of data flexibility. Numerous studies have explored GNN-based Person Re-ID methods, including those using Graph Convolutional Networks (GCN) and methods based on graph self-attention mechanisms. Liu et al. \cite{liuDeepAdversarialGraph2019} proposed a novel deep adversarial graph attention convolutional network (A-GANet) for text-based person search. A-GANet utilizes text and visual scene graphs, including object properties and relationships, from pedestrian text queries and gallery images to learn information-rich text and visual representations. It learns an effective joint text-visual latent feature space through adversarial learning, bridging the modality gap and promoting person matching. Li et al. \cite{liPersonTextImageMatching2022} presented a fundamental framework using graph convolution as a feature representation. They simulate adversarial attacks caused by text and image diversity on feature extraction by embedding additional attack nodes in graph convolution layers to enhance the model's robustness to textual and visual diversity. However, when graph neural networks compute edges for pedestrian body nodes and entity relationships, it may result in high computational complexity and memory requirements. Moreover, GNNs are more susceptible to overfitting compared to general network structures.

\subsection{Transformer}
Due to the success of Transformers in multimodal tasks, it has become a popular architecture widely employed in text-based Person Re-ID tasks \cite{chenTIPCBSimpleEffective2022,liTransformerBasedLanguagePersonSearch2021,gaoConditionalFeatureLearning2022,shaoLearningGranularityUnifiedRepresentations2022,jiAsymmetricCrossScaleAlignment2022,liJointTokenFeature2022a,yanCLIPDrivenFinegrainedTextImage2022,wangExploitingTextualPotential2023,yangUnifiedTextbasedPerson2023a}. Currently, a commonly used approach involves a dual-stream network architecture, where one stream is dedicated to processing pedestrian images, and the other handles the corresponding text descriptions. In the image stream, convolutional neural networks \cite{zhuDSSLDeepSurroundingsperson2021} or vision transformers \cite{shuSeeFinerSee2023} are utilized to extract visual features from pedestrian images. Simultaneously, in the text stream, Long Short-Term Memory (LSTM) \cite{wuLapsCoreLanguageGuidedPerson2021} or transformers \cite{gaoConditionalFeatureLearning2022} are applied to extract feature vectors from the corresponding text descriptions. Li et al. \cite{liTransformerBasedLanguagePersonSearch2021} introduced a transformer-based Person Re-ID architecture, evaluating the shared attention between language reference terms and visual features through transformer blocks. Apart from achieving outstanding search performance, the proposed method also provides interpretability by visualizing the attention between image parts in person pictures and their corresponding referential terms. Existing works often overlook the granularity differences between features of the two modalities, where visual features tend to be fine-grained while text features are coarse-grained, leading to significant modality gaps. Shao et al. \cite{shaoLearningGranularityUnifiedRepresentations2022} proposed an end-to-end framework named Learning Granularity-Unified Representations (LGUR), based on the transformer, aiming to map visual and text features to a unified granularity space. Addressing alignment deficiencies between image region features and text features in weakly supervised learning methods, resulting in poor feature distribution, Gao et al. \cite{gaoConditionalFeatureLearning2022} presented a novel method called Conditional Feature Learning based Transformer (CFLT). CFLT maps sub-regions and phrases to a unified latent space and dynamically adjusts features from one modality based on data from another modality by constructing explicit conditional embeddings.

With the significant progress achieved by large models, Sarafianos et al. \cite{sarafianosAdversarialRepresentationLearning2019} demonstrated for the first time the effectiveness of BERT as an openly accessible word embedding extraction language model in the domain of image-text matching. Subsequently, they combined two feature vectors using a cross-attention mechanism and applied them to pedestrian re-identification tasks. Li et al. [70] proposed a semantically aligned feature aggregation network that adaptively aggregates unit features with the same semantics into different part-perceptive features. Initially introducing Transformer-based backbones ViT \cite{dosovitskiyImageWorth16x162023} and BERT into visual and text modalities, Ji et al. \cite{liPersonTextImageMatching2022} extracted robust feature representations for text-based person search. They proposed an Asymmetric Cross-Scale Alignment (ACSA) method, employing Swin Transformer for visual representation extraction and BERT for text and phrase representation in the text domain. In contrast to previous methods that use two separate networks for feature extraction, Wang et al. \cite{wangCAIBCCapturingAllround2022} introduced an Implicit Visual-Text (IVT) framework, a single-stream network architecture using only a single transformer. Text and image feature extraction utilizes a backbone network, optimizing network shared parameters with cross-modal data to facilitate learning a common space mapping. Recently, Vision-Language Pre-training (VLP) models in visual language pre-training have demonstrated the transferability of knowledge for downstream Text-Picture Search (TPS) tasks, enhancing performance effectively. An approach called CLIP-driven Fine-grained Information Mining (CFine) \cite{wangExploitingTextualPotential2023} was proposed to leverage CLIP's visual knowledge for person re-identification, incorporating fine-grained information mining to explore intra-modal discriminative clues and inter-modal correspondences. However, the lack of VLP text representations may compromise the valuable cross-modal alignment of the two encoders. Hence, Wang et al. \cite{wangExploitingTextualPotential2023} introduced a model with two pre-training modes, TP-TPS, where both text and image encoders are based on transformers, and all backbones are initialized with pre-training weights from CLIP. They introduced Multi-Integrity Description Constraints (MIDC) to enhance the robustness of the text modality by combining different components of a fine-grained corpus during training. The advantage of the Transformer model lies in its ability to effectively learn sequence information, making it capable of capturing information in text descriptions efficiently. Additionally, this approach exhibits good interpretability in the field of multi-scale learning, as it allows the visualization of regions of interest in pedestrian images and keywords in text descriptions based on attention weights. Finally, it plays a crucial role in unified modality architectures.


\section{Optimization}
\begin{figure*}[!t]
  \centering
  \includegraphics*[width=\textwidth]{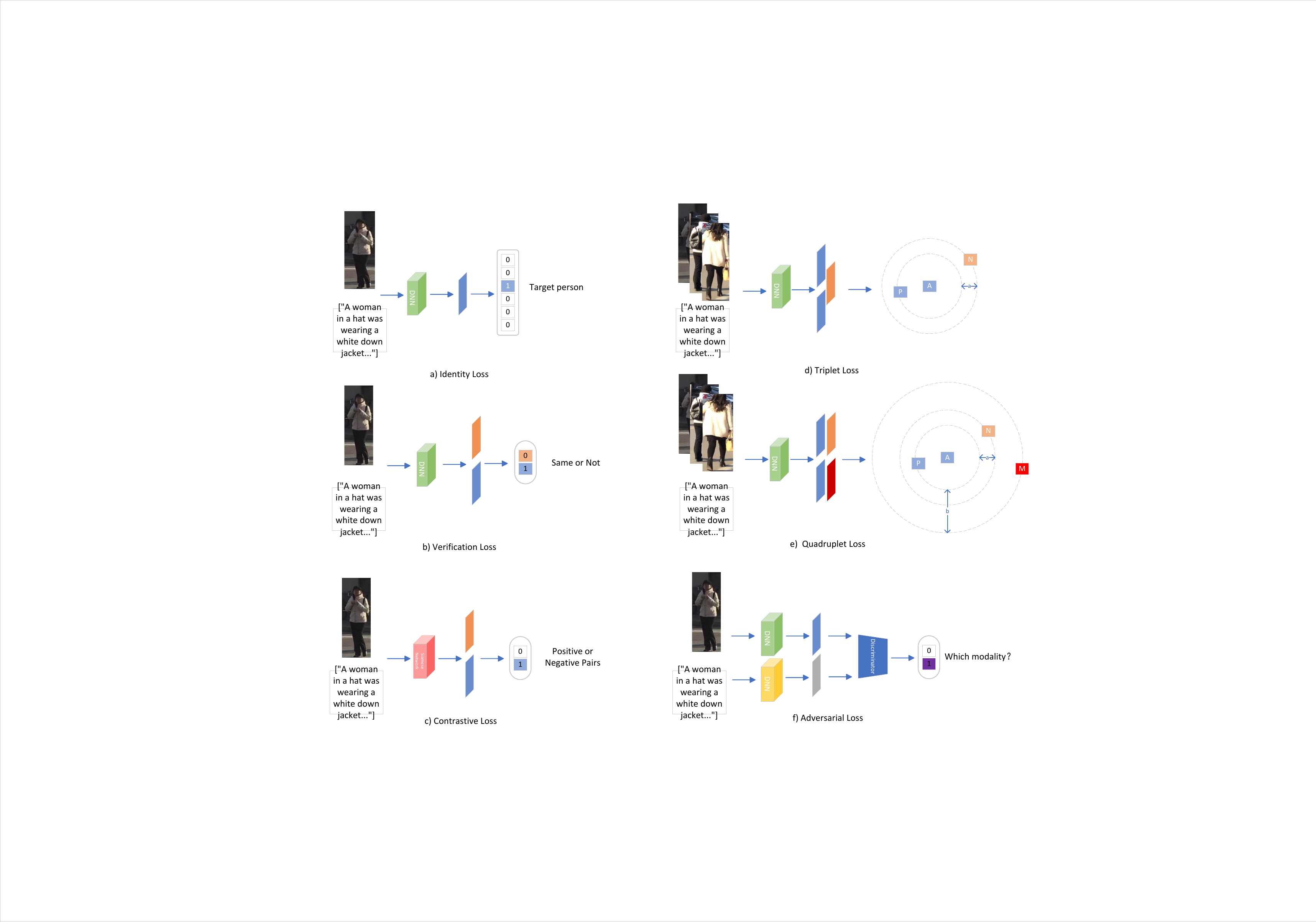}
  \caption{The figure shows the principles of the most prominent alignment loss methods in text-based person re-identification. These include identity loss, verification loss, comparison loss, triplet loss, quadruplet loss and adversarial loss. The vast majority of new alignment methods being devised are currently based on variants or combinations of these losses.}
  \label{fig4}
\end{figure*}

Deep metric learning plays a paramount optimization role in cross-modal person re-identification \cite{zhengDiscriminativelyLearnedCNN2018}. 
For instance, it enhances the discriminative power of feature representation, addresses a multitude of noise and variations such as diverse lighting conditions, poses, and occlusions, adapts the distance metric accordingly, embeds more semantically meaningful spatial structures, and leverages domain adaptation to tackle the challenges posed by multiple viewing angles. Of particular significance is the optimization of loss functions, thereby improving the matching performance in cross-modal person re-identification tasks \cite{duanDeepLocalizedMetric2018}. This approach effectively addresses a spectrum of challenges, elevating the network's robustness and generalization capacity, consequently achieving more accurate person matching across distinct camera scenes and conditions. 
This study places emphasis on prevalent loss functions used in text-based person re-identification tasks, including identity loss, verification loss, contrastive loss, triplet loss, quadruplet loss, and adversarial loss, as shown in Figure~\ref{fig4}.

\subsection{Identity Loss}
Zheng et al. \cite{zhengPersonReidentificationPresent2016} introduced the ID Discriminative Embedding (IDE) network, which treats each pedestrian as a separate category and employs the pedestrian's ID as a classification label to train a deep neural network. This allows the network to learn the capability of predicting the identity given a pedestrian image. The classification loss is typically computed using the cross-entropy loss function. During training, the classifier optimizes its weight parameters by minimizing the classification loss. A fully connected layer (FC) for classification is appended to the end of the network, and the softmax activation function maps the features to a probability space representing identity labels \cite{geVisualTextualAssociationHardest2019a}. The cross-entropy loss for the multi-class pedestrian Re-ID task can be expressed as:

\begin{equation}
L_{ID} = -\frac{1}{N}\sum_{i=1}^{N}y_{i}\log(p_{i}).
\end{equation}

Here, $x_{i}$ denotes the input image, $y_{i}$ represents the corresponding label, $p_{i}$ is the probability that $x_{i}$ is recognized as $y_{i}$ after the softmax function, and $N$ indicates the batch size. Identity loss is widely adopted in pedestrian re-identification \cite{wangCAIBCCapturingAllround2022,farooqCrossModalPerson2020,dosovitskiyImageWorth16x162023,sarafianosAdversarialRepresentationLearning2019, wangAMENAdversarialMultispace2021,suoSimpleRobustCorrelation2022a,farooqConvolutionalBaselinePerson2020,luoSpectralFeatureTransformation2018,wangSUMSerializedUpdating2022}, due to its automatic hard sample mining and ease of training. However, as the labeled data increases significantly, issues related to decreased efficiency in classifier training might arise, especially when dealing with datasets containing a large number of similar pedestrians. In cross-modal pedestrian re-identification, the classification loss is often used in conjunction with other loss functions \cite{zhaoWeaklySupervisedTextBased2021, hanTextBasedPersonSearch2021a}, such as contrastive loss and triplet loss \cite{zhengHierarchicalGumbelAttention2020,liIdentityAwareTextualVisualMatching2017,yanImageSpecificInformationSuppression2022,wangCAIBCCapturingAllround2022}, to enhance performance of the model.

\subsection{Verification Loss}
The verification loss is employed to quantify the similarity between two pedestrian images \cite{zhengReIdentificationConsistentAttentive2019}. In pedestrian re-identification, it is often necessary to compare two images to determine if they belong to the same individual. The verification loss aims to achieve this objective by comparing the feature vectors of two images to compute their similarity \cite{zhengDiscriminativelyLearnedCNN2018}. Specifically, the verification loss is calculated by inputting the feature vectors of two images into a similarity metric function. This similarity metric function can be Euclidean distance, cosine similarity, or other similar measures. If two images belong to the same person, their feature vectors should exhibit high similarity, resulting in a high similarity score \cite{liDeepReIDDeepFilter2014}. Thus, the verification loss is minimized to enhance the similarity between two images of the same person. The specific formula is as follows:

\begin{equation}
L_{veri} = -\frac{1}{N}\sum_{i=1}^{N}y_{i}\log(p_{i}) + (1-y_{i})\log(1-p_{i}).
\end{equation}

Here, $N$ is the number of sample pairs, $y_{i}$ is a label indicating whether the sample pair $x_{i}$ and $x_{j}$ belongs to the same category ($1$ denotes the same category, $0$ denotes different categories), and the softmax function is utilized to calculate the probability $p_{i}$ of having the same label for both samples. Due to the necessity of pairwise computation, the verification loss tends to have lower efficiency in re-identification tasks. Therefore, it is often used in combination with other losses, such as identity loss, to leverage the advantages of both methods, thereby enhancing recognition speed and accuracy.

\subsection{Contrastive Loss}
Contrastive Loss was originally introduced by Sumit Chopra et al. in the context of face verification tasks \cite{chopraLearningSimilarityMetric2005}, aiming to train neural networks to determine whether two inputs belong to the same facial identity. In the context of person re-identification, the principle of encouraging similar identity feature representations to be closer in the feature space while dispersing feature representations of different identities is leveraged for model training within Siamese networks \cite{wangPersonReIdentificationCascaded2018}. The Contrastive Loss not only considers the distances between images of the same identity but also takes into account the distances between images of different identities. The function can be expressed as follows:

\begin{equation}
L_{con} = \frac{1}{2N}\sum_{n=1}^{N}y_{n}D_{n}^{2} + (1-y_{n})\max(margin - D_{n}, 0)^{2}.
\end{equation}

Here, $N$ is the number of sample pairs, $y_{n}$ is a label indicating whether sample pairs $x_{i}$ and $x_{j}$ belong to the same class (1 for the same class, 0 for different classes), $D_{n}$ represents the distance between feature representations (Euclidean distance, cosine distance, etc.), and $margin$ is a threshold indicating that when sample pairs belong to different classes, their distance should be greater than this value. For sample pairs belonging to the same class, the loss is the square of the distance $D_{n}$, while for sample pairs belonging to different classes, the loss is the square of the difference between the distance and the threshold value $margin$. This framework aims to bring feature representations of sample pairs from the same class closer together and push feature representations of sample pairs from different classes further apart. By adjusting the value of $margin$ and the weight of the loss, the distance range between sample pairs can be controlled, influencing the learning of feature representations \cite{variorSiameseLongShortTerm2016}.

\subsection{Triplet Loss}
The Triplet Loss is a widely utilized loss function in pedestrian re-identification, aiming to learn discriminative feature representations by comparing the distances between three images \cite{wangTextbasedPersonSearch2021,wangCAIBCCapturingAllround2022,chechikLargeScaleOnline2010,hermansDefenseTripletLoss2017,wangAMENAdversarialMultispace2021,farooqAXMNetImplicitCrossModal2022,yanImageSpecificInformationSuppression2022,wangSUMSerializedUpdating2022,jingPoseGuidedJointGlobal2018,iodiceTextAttributeAggregation2020,wangCAIBCCapturingAllround2022}. Its fundamental concept involves learning feature representations with discrimination, such that the distance between images of the same person is minimized, while the distance between images of different individuals is maximized \cite{CascadeAttentionNetwork2018}, \cite{geVisualTextualAssociationHardest2019a}, achieved by minimizing the Triplet Loss during training. Specifically, given an anchor image, a positive sample image, and a negative sample image, the objective of Triplet Loss is to minimize the distance between the anchor and the positive sample while simultaneously maximizing the distance between the anchor and the negative sample \cite{zhengHierarchicalGumbelAttention2020}, \cite{wangLanguagePersonSearch2019}. The formal definition of Triplet Loss is as follows:

\begin{equation}
\begin{aligned}
L_{triplet} = & \sum_{n=1}^{N}\max(||f(x_{n}^{a}) - f(x_{n}^{p})||_{2}^{2}\\
              & \quad - ||f(x_{n}^{a}) - f(x_{n}^{n})||_{2}^{2} + margin, 0).
\end{aligned}
\end{equation}

Here, $f$ is a function that maps an image $x$ to a feature vector, $x_{n}^{a}$, $x_{n}^{p}$ and $x_{n}^{n}$ are the anchor, positive, and negative sample images of the $n$-th instance, respectively. $||\cdot||_{2}$ represents the Euclidean distance between two embedding vectors, $\max(\cdot)$ represents taking the maximum value between $0$ and $\cdot$, $margin$ is a hyperparameter controlling the difference in distances between images of the same person and images of different people. It is worth noting that in practice, Triplet Loss has encountered certain issues, such as sample selection and slow learning speed. Consequently, improved versions have been proposed \cite{liIdentityAwareTextualVisualMatching2017,farooqAXMNetImplicitCrossModal2022,wangCAIBCCapturingAllround2022}. Wang et al. \cite{wangLanguagePersonSearch2019} proposed a novel mutually connected classification loss (MCCL) based on triplet loss to fully exploit identity-level information. It not only incorporates identification information into both images and language descriptions but also encourages similar cross-modal classification probabilities for instances belonging to the same identity. These improved versions prove to address certain existing issues more effectively and have demonstrated excellent performance in person re-identification tasks.

\subsection{Quadruplet Loss}
The Quadruplet Loss is an extension of the Triplet Loss \cite{chenTripletLossDeep2017}. The Triplet Loss is commonly employed to train a neural network to embed images of the same person into close proximity within the embedding space, while images of different individuals are embedded far from the space occupied by the same person. However, the Triplet Loss may suffer from issues such as gradient vanishing or exploding, which could render it less stable in practical usage. The Quadruplet Loss enhances the stability of the Triplet Loss by introducing a fourth sample.

In specific terms, the Quadruplet Loss computes pairs of embedding distances for two images of the same person and two images of different individuals, and employs these four distances to formulate the loss function. In this regard, the two images of the same person should be closer, while the two images of different individuals should be farther apart. By separately using the two images of the same person and the two images of different individuals to calculate distances, the Quadruplet Loss can yield more stable training, thus enhancing the performance of person re-identification. The formula for the Quadruplet Loss is as follows:

\begin{equation}
\begin{aligned}
    L_{quad} = \sum_{n=1}^{N}\max(&||f(x_{n}^{a}) - f(x_{n}^{p})||_{2}^{2} - ||f(x_{n}^{a}) - f(x_{n}^{n1})||_{2}^{2}\\
    & + \alpha, 0)+ \max(||f(x_{n}^{a}) - f(x_{n}^{n1})||_{2}^{2}\\
    &- ||f(x_{n}^{n1}) - f(x_{n}^{n2})||_{2}^{2} + \beta, 0).
\end{aligned}
\end{equation}

Here, $f$ denotes the embedding function of the neural network, $x_{n}^{a}$, $x_{n}^{p}$, $x_{n}^{n1}$ and $x_{n}^{n2}$ respectively represent the anchor image, positive sample image, first negative sample image, and second negative sample image for the $n$-th sample. $||\cdot||_{2}$ represents the Euclidean distance between two embedding vectors. $\max(\cdot)$ represents taking the maximum value between $0$ and $\cdot$, The hyperparameters $\alpha$ and $\beta$ control the difference in distances between the two images of the same person and the two images of different individuals.

The Quadruplet Loss effectively reduces the within-class variance and improves the discriminative performance of the model by introducing an additional negative sample. However, the increase in parameters also requires the computation of more pairs of samples, leading to a significant increase in computational complexity and longer training time. Therefore, suitable strategies are required to select effective quadruples in practical applications.

\subsection{Adversarial Loss}
The adversarial loss primarily leverages adversarial training to learn more robust pedestrian feature representations \cite{liPersonTextImageMatching2022}. Its fundamental concept involves mapping pedestrian feature embeddings from the training data into a similarity space, where the distribution of generated image features is made closer to that of real image features, enhancing the robustness of the network in handling multi-modal information \cite{sarafianosAdversarialRepresentationLearning2019}. Specifically, adversarial loss employs adversarial samples during training, where the model seeks to minimize the adversarial loss. The computation of the adversarial loss involves introducing perturbations to the model input, making the input pedestrian image appear visually similar to the original image while introducing significant disparities in the embedding space. These perturbations are generated by a generator network, which attempts to produce images similar in appearance to the input image but with substantial deviations in the embedding space. The discriminator seeks to maximize the adversarial loss to effectively discriminate between these two types of features. The general formula for adversarial loss is as follows:

\begin{equation}
\begin{aligned}
L_{adv} = &\mathbb{E}_{x\sim p_{data}(x)}[\log D(x)] \\
&+ \mathbb{E}_{z\sim p_{z}(z)}[\log(1 - D(G(z)))].
\end{aligned}
\end{equation}

Here, $G$ is the generator, $D$ is the discriminator, $x$ represents real image features, $z$ represents generated image features, $\log D(x)$ denotes the logarithm of the discriminator's predicted probability for real image features, and $\log(1 - D(G(z)))$ denotes the logarithm of the discriminator's predicted probability for generated image features. Liu et al. \cite{liuDeepAdversarialGraph2019} proposed the application of adversarial principles in pedestrian re-identification to acquire more robust feature representations. They contend that the distributions of text and image features should be modality-invariant and semantically discriminative, while maintaining latent cross-modal semantic similarities. These features exhibit enhanced resistance to adversarial attacks. Subsequently, an adversarial learning module was devised to optimize the representations extracted from image graph attention networks and text graph attention networks. The objective of the modality discriminator is to reliably distinguish between modalities of samples as much as possible, given a well-trained feature transformer, until the point where discrimination becomes infeasible. Adversarial loss is employed to classify the modality labels (belonging to either image or text) of input samples. Despite its relatively recent application in pedestrian re-identification, adversarial loss has demonstrated significant efficacy in some recent studies \cite{wangAMENAdversarialMultispace2021}.


\section{Challenge}

In this paper, we first give an overview of text-based person Re-ID in two parts: attribute-based and natural language-based person Re-ID. Secondly, the datasets based on the two parts are summarized and described respectively. Moreover, the commonly used technical strategies in text-based person Re-ID are analyzed. In addition, an overview of the deep learning architectures that usually appear in text-based person Re-ID is made. Further, the loss functions (alignment optimization methods) commonly used in text-based person Re-ID are summarized and described in detail. Finally, we analyze the challenges and possible future directions of text-based person Re-ID from the perspectives of feature extraction, data scarcity, pre-trained models, and generative models, respectively.

Although text-based person Re-ID technology offers numerous advantages in fields such as video surveillance, security, social media, and virtual reality\cite{gaoContextualNonLocalAlignment2021}, it still faces various challenges. The most critical issues include the ambiguity and granularity of natural language. Identical descriptions may be incomplete, ambiguous, or contain ambiguities, potentially leading to incorrect identity matches. On the other hand, photographs available in real-world galleries pertaining to pedestrians often lack high resolution, ample lighting, and freedom from obstruction\cite{farooqConvolutionalBaselinePerson2020}. 
Zhang et al. \cite{zhang2022realgait} demonstrated recently that gait information may be the basis for general person Re-ID, and that its features may be more robust compared to the highly homogeneous pedestrian appearance feature signature. However, existing gait recognition methods are not able to meet the requirements of less controlled scenarios, e.g., most of the studies are based on well controlled scenarios such as narrow corridors. If one wants to improve the efficiency of person Re-ID by combining gait features with appearance features, it may be necessary to obtain more coherent and high-quality gait data, while the appearance is recognizable. 
Therefore, integrating text descriptions with image/video information requires more efficient and robust cross-modal fusion techniques for the system to better understand and match entities. Pre-trained large-scale multimodal models may represent a reasonable direction for achieving improved encoding effects. Furthermore, text-based person re-identification necessitates more high-quality data to enable the model to learn relationships between various descriptions and identities \cite{yangUnifiedTextbasedPerson2023a}. It is noteworthy that the recent popularity of text-to-image generation models, as a novel approach in pedestrian re-identification, holds significant potential, as these models generate graphics from textual descriptions for retrieval. Given that cameras in the real world do not always capture images from an available query library, customizing the generation of corresponding target entities based on eyewitness text is meaningful. However, due to the insufficient maturity of text-to-image generation model controllability, we remain optimistic about this direction.


\section{Foresight}

\subsection{\bf A Novel Baseline}
Since the current text-based person Re-ID tasks being based on disparate datasets, specialized models are separately trained on large samples. However, in the real world, the lack of a uniform style (resolution, size, angle, brightness) in pedestrian Re-ID data is often due to the constraints imposed by different specifications of data collection devices (cameras). Retrieving arbitrary pedestrian targets in an open-world scenario becomes quite challenging for dedicated models. There is an urgent need for a new approach that enables zero-shot retrieval of targets from any gallery. On the other hand, text-to-image retrieval fundamentally relies on comparing cross-modal data distributions from small datasets through contrastive learning. If the gallery itself lacks target pedestrians, this results in ineffective retrieval, contradicting the concept of open-world pedestrian recognition. Inspired by the controllable generation retrieval in text-to-image for pedestrians, we propose a novel baseline model for text-based person re-identification, which involves re-retrieving target images from text to image.

\begin{figure}[!h]
  \centering
  \includegraphics[width=3.6in]{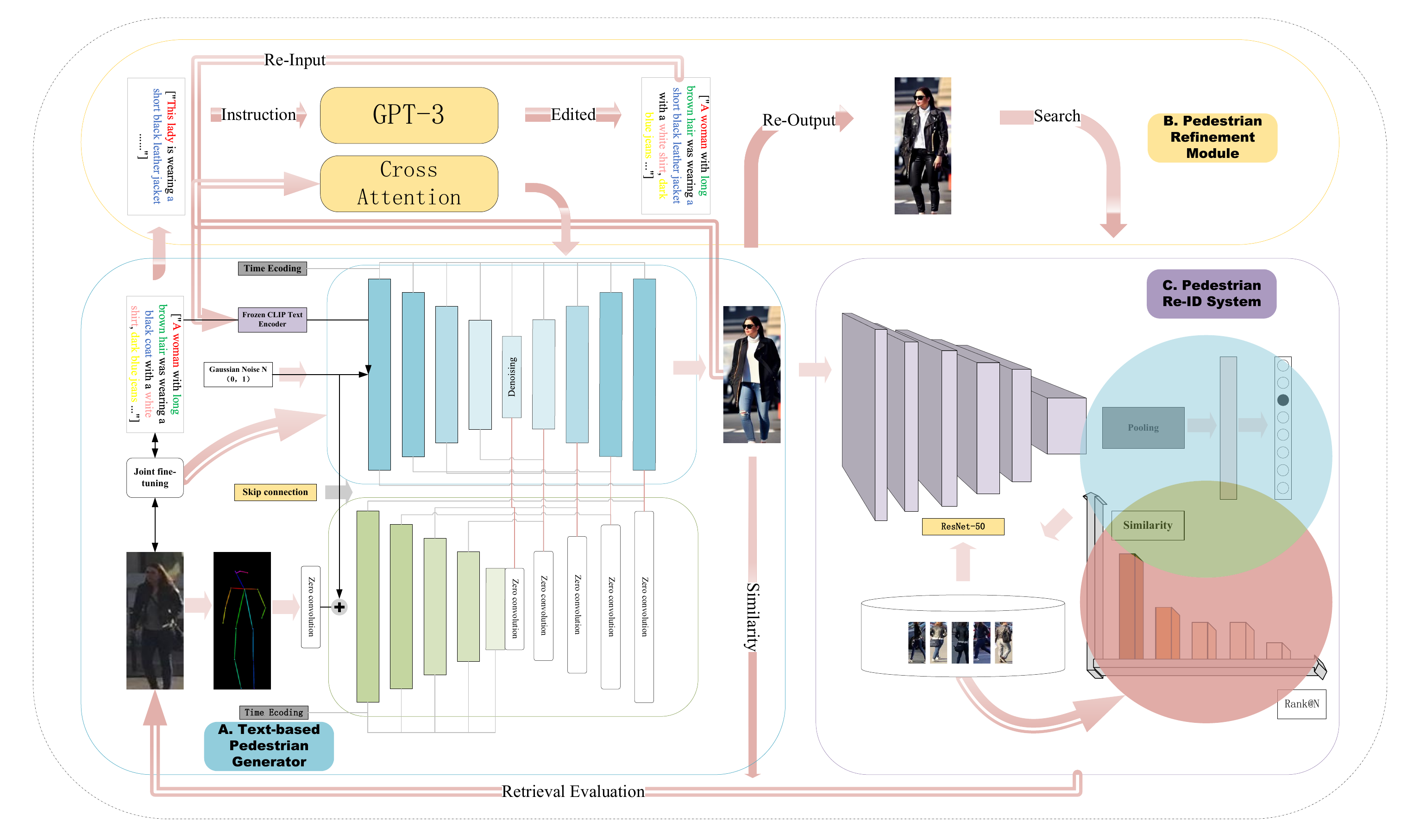}
  \caption{Text-Based Pedestrian Image Generation Guided Re-ID baseline architecture. The entire architecture consists of three modules: A) Text-based pedestrian generator, B) Pedestrian refinement module, and C) Pedestrian re-recognition system. The entire architectural flow can be viewed starting with module A at the bottom left, preferably in a clockwise direction. Note that there may not be some kind of very clear boundary between modules A and B, and that there is interaction between them in the pedestrian generation process.}
  \label{fig6}
\end{figure}

As illustrated in Figure \ref{fig6}, the entire text-based pedestrian image generation guided Re-ID baseline architecture is demonstrated. The initial step involves encoding natural language descriptions of pedestrians and utilizing a diffusion model to generate corresponding pedestrian images. Subsequently, these generated pedestrian images are employed in the task of pedestrian re-identification. Given that current text-to-image generation models primarily focus on coarse-grained overall image generation tasks, emphasizing inter-class coordination among different objects rather than concentrating on intra-class distinctiveness in human images, and while simultaneously considering authenticity, a strategy based on human body key-point assistance is adopted. This strategy involves extracting features from human body key-points to assist in controlling the pedestrian generation process, ensuring compliance with the requirement for the presence of the majority of the human body in the pedestrian re-identification task. The generated pedestrian images serve as queries for retrieval and are input into the backbone network for visual feature extraction. To enhance retrieval accuracy, emphasis is placed on restoring features with the highest relevance to the original image. For instance, in comparison to generated images, sensitivity to channel attention may be more pronounced in original images. Various attention mechanisms are employed to assess their impact on retrieval effectiveness. However, due to the coarser granularity of linguistic descriptions compared to images, the generation of pedestrian images may not be sufficiently detailed in terms of the style and material of the pedestrian's clothing. For example, the textual description may be as broad as "black coat", but the image may be as specific as "short black leather jacket", which results in feature learning inaccuracy. In order to improve the accuracy of localized feature generation for pedestrians in a realistic open set, an approach using cross-attention based image editing is recommended. After using natural language instructions, the local features of pedestrians are controlled to change, and the witness's memory of the local features for the target person is refined and specialized. Finally, the results of pedestrian retrieval are obtained through different optimization strategies.\\

\begin{table*}[!t]
  \centering
  \caption{The table demonstrates the results of person Re-ID with and without fine-tuning the stable diffusion model. Tests are performed on three datasets, CUHK-PEDES, RSTPReID and ICFG-PEDES, respectively, using different visual representation backbone networks.}
  \label{experiment results}
  \begin{tabular}{c|c|c|c|c|c|c|c|c|c|c}
  \hline
\multirow{2}{*}{\textbf{Stable Diffusion}}  & \textbf{Method} & \multicolumn{3}{c}{\textbf{CUHK-PEDES}} & \multicolumn{3}{c}{\textbf{RSTPReID}} & \multicolumn{3}{c}{\textbf{ICFG-PEDES}}\\
               & & rank@1 & rank@5 & rank@10 & rank@1 & rank@5 & rank@10 & rank@1 & rank@5 & rank@10 \\
  \hline
  \hline
  \multirow{4}{*}{Non-Finetuned}
  &DenseNet               & 0.0396 & 0.1072 & 0.1944 & 0.0275 & 0.0980 & 0.1882 & 0.0077 & 0.0210 & 0.0323\\
  &EfficientNet           & 0.0584 & 0.1233 & 0.2216 & 0.0392 & 0.1255 & 0.2000 & 0.0087 & 0.0209 & 0.0326\\
  &Swin Transformer       & 0.0630 & 0.1519 & 0.2460 & 0.0471 & 0.1294 & 0.2235 & 0.0084 & 0.0215 & 0.0326\\
  &ResNet                 & 0.0674 & 0.1794 & 0.2578 & 0.0491 & 0.1353 & 0.2348 & 0.0103 & 0.0287 & 0.0392\\
  \hline
  \hline
  \multirow{4}{*}{Finetuned}
  &DenseNet               & 0.1101 & 0.2179  & 0.3298 & 0.0745 & 0.2196 & 0.3176 & 0.0328 & 0.0783 & 0.1278 \\
  &EfficientNet           & 0.1498 & 0.2619  & 0.4305 & 0.1103 & 0.2784 & 0.3686 & 0.0651 & 0.1254 & 0.1701 \\
  &Swin Transformer       & \textbf{0.1622} & \textbf{0.3794}  & 0.4675 & 0.1098 & 0.2667 & 0.3804 & 0.0632 & 0.1302 & 0.2015 \\
  &ResNet                 & 0.1586 & 0.3648  & \textbf{0.4827} & \textbf{0.1176} & \textbf{0.2824} & \textbf{0.4118} & \textbf{0.0749} & \textbf{0.1306} & \textbf{0.2191} \\
  \hline
  \end{tabular}
\end{table*}

\subsection{\bf Experiment \& Results}
\textbf{For A) Text-based pedestrian generation module}, We employed the Frozen CLIP Text Encoder \cite{radfordLearningTransferableVisual2021} to encode pedestrian descriptions, followed by inputting them into stable diffusion \cite{rombachHighResolutionImageSynthesis2022} to generate pedestrian images. The denoising module utilized the cross-attention U-Net \cite{ronnebergerUNetConvolutionalNetworks2015} backbone, which proves beneficial for the generation of high-quality pedestrian images. For pedestrian pose assistance, we adopt ControlNet \cite{zhangadding2023} for Pedestrian integrity control, which ensures that the generated pedestrian image has full body. Specifically, OpenPose \cite{caoOpenPoseRealtimeMultiPerson2019} was employed for detecting key points on the pedestrian body. Subsequently, pose features were extracted through skip connections and zero convolutions, facilitating their incorporation into pedestrian generation. 

\textbf{In the part of B) pedestrian local feature refinement}, we input the local image editing instructions and then use GPT-3 \cite{brown2020language} to rewrite and replace the pedestrian image generation instructions. Re-inputting the rewritten text cue and the pedestrian image before rewriting into the stable diffusion model, the visual and textual features of pedestrians are fused using the cross-attention layer, and the image can be edited by editing the text only to realize the mapping from the source image to the target image \cite{hertz2022prompt}. As a comparative analysis, we also attempted fine-tuning the baseline model using a subset of text-image pairs data based on text-driven pedestrian re-identification. In the retrieval module, the generated pedestrian images were employed as queries. 

\textbf{In the C) pedestrian Re-ID system}, ResNet-50 \cite{heDeepResidualLearning2016} served as the visual backbone, incorporating self-attention mechanisms. The Part-based Convolutional Baseline (PCB)\cite{sunPartModelsPerson2018} strategy was employed for segmenting human bodies to extract feature maps, and random horizontal image flipping was applied for data augmentation. The training process involved the use of Instance Batch Normalization (IBN) \cite{panTwoOnceEnhancing2020}strategy, with a batch size set to 32, droprate at 0.5, erasing probability at 0.2, ins-gamma set to 32, learning rate at 0.05, and weight decay at 0.0005. Furthermore, the training amalgamated ID loss, Contrastive loss, and Triplet loss as the loss functions. The optimization utilized SGD optimizera with momentum, and the model was trained for 60 epochs.

\begin{figure*}[!t]
  \centering
  \includegraphics*[width=\textwidth]{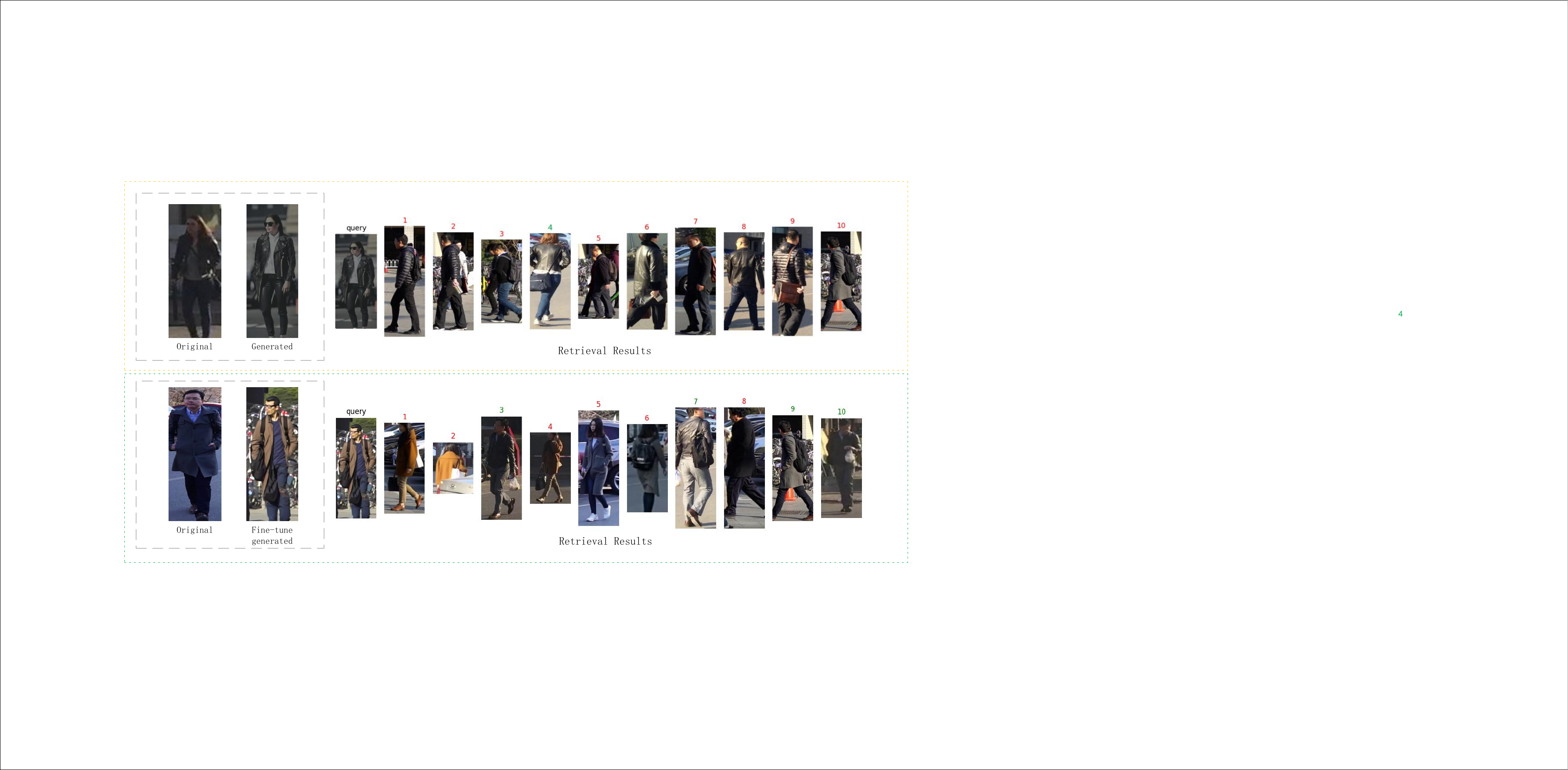}
  \caption{Generated images of text-based pedestrians without and with fine-tuning are shown, along with their retrieval Rank@10 results. The green numbers shown at the top of the image represent a positive retrieval and the red color represents a negative retrieval.}
  \label{fig8}
\end{figure*}

\begin{figure*}[!t]
    \centering
    \includegraphics*[width=\textwidth]{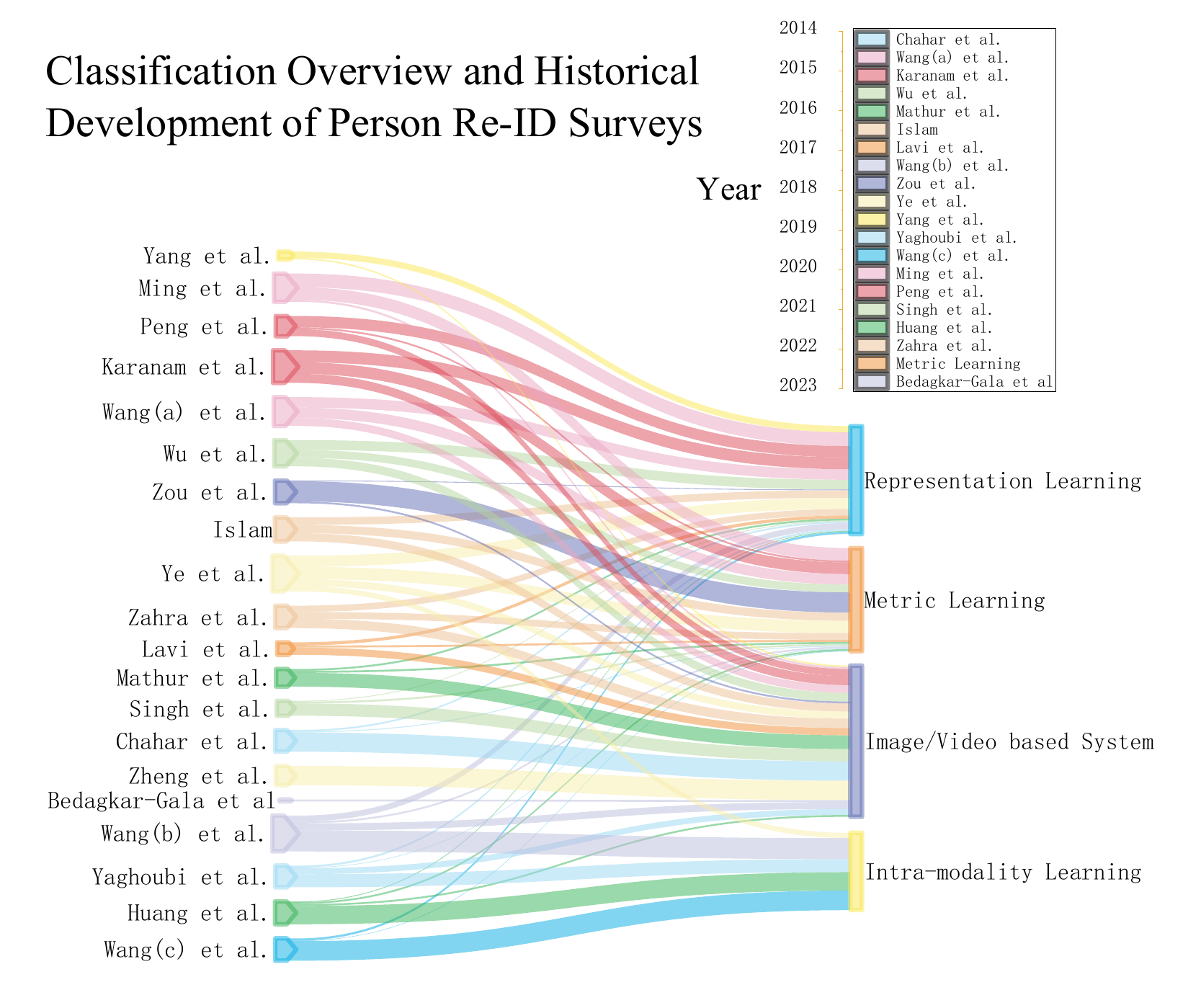}
    \caption{The figure shows the development of the main categorical dimensions and surveys of Person Re-ID throughout history. Overall, the large figure on the left represents the relevance of influential Person Re-ID surveys and their categorical dimensions, with the line font implicitly indicating their phase weights. The smaller figure on the top right is a more visual representation of the temporal relationships between the emergence of these Person Re-ID surveys.}
    \label{fig2}
  \end{figure*}

As depicted in the Table ~\Ref{experiment results}, we evaluated the proposed Text-Based Pedestrian Image Generation Guided Re-ID (TBPGR) baseline architecture on three widely used text-based person Re-ID datasets (CUHK-PEDES, RSTPReid, ICFG-PEDES). Specifically, retrieval of pedestrian images was generated using the generation modules with and without fine-tuned Stable Diffusion, respectively. We report the performance of visual representation modules DenseNet, EfficientNet Swin-Transformer and ResNet on the task for each dataset and each condition of with and without fine tuning. 
ResNet generally shows better performance. After the image generation in the intermediate step, we qualitatively analyzed the content of the images. It was found that the quality of the pedestrian images generated before fine-tuning was higher, and the quality of the images generated after fine-tuning became lower, but the retrieval efficiency was slightly improved, conjecturing that the quality of the pedestrian image text data pairs had some effect on the data distribution. 
Besides, since the CUHK-PEDES dataset retrieval effect is better than other datasets, it is hypothesized that it is because the generation of pedestrian images seldom involves the factors such as lens occlusion and illumination changes, which have some implications on the pedestrian retrieval performance, as shown in Figure ~\ref{fig8}.

Although our baseline model(TBPGR) is still not powerful enough in comparison to traditional training approaches based on implicit spatial alignment of text and images, we believe that such an open-retrieval paradigm is far more valuable in future Open Set text-based person Re-ID scenarios than just pursuing the accuracy of the present moment.

\section*{Acknowledgments}
The authors would like to thank all the anonymous reviewers and associate editors for their valuable comments to improve the paper. The authors would also like to thank the authors of the three publicly available datasets.


 

\section*{APPENDIX}
This appendix provides tables and figures that supplement the data and analyses presented in the main body of the paper. 
These tables and figures provide more detailed information and further support the findings discussed. 
They are referenced throughout the paper where relevant.

\begin{table*}[t]
  \caption{The table compares the aspects emphasized by existing surveys based on chronological order in terms of the dimensions of authors, journal name, and major contributions.}
  \label{table1}
    \begin{tabularx}{\textwidth}{|p{3cm}|X|p{2cm}|X|}
    \hline
    \textbf{Author} & \textbf{Publication} & \textbf{Reference} & \textbf{Contribution} \\
    \hline
    Bedagkar-Gala et al. &A survey of approaches and trends in person re-identification \cite{bedagkar-galaSurveyApproachesTrends2014b} & IVC14 & Introduced person Re-ID challenges up to 2014 and summarizes the mainstream solutions.\\
    \hline
    Zheng et al. & Person Re-identification Past, Present and Future \cite{zhengPersonReidentificationPresent2016} & arXiv16 & Discussed the past development mainly in terms of pedestrian retrieval, and image classification. They investigated the image/video-based systems and methods.\\
    \hline
    Chahar et al. & A study on deep convolutional neural network based approaches for person re-identification \cite{chaharStudyDeepConvolutional2017} & PRMI17 & Presented the methods for pedestrian Re-ID based on CNNs and explained some of the problems and directions for development.\\
    \hline
    Wang et al. & Survey on person re-identification based on deep learning \cite{wangSurveyPersonReidentification2018}& CAAI18 & Surveyed deep learning-based pedestrian re-recognition approaches, and pointed out the ways forward at that time.\\
    \hline
    Karanam et al. & A systematic evaluation and benchmark for person re-identification: Features, metrics, and datasets \cite{karanamSystematicEvaluationBenchmark2019} & TPAMI19 & Conducted a comprehensive review and performance evaluation of single- and multi-shot Re-ID algorithms. Person Re-ID was first started to be introduced in terms of feature extraction and metric learning.\\
    \hline
    Wu et al. & Deep learning-based methods for person re-identification: A comprehensive review \cite{wuDeepLearningbasedMethods2019} & NC19 & Summarized six deep learning-based methods for person re-identification, including recognition, verification, distance metrics, parts, video, and data augmentation.\\
    \hline
    Mathur et al. & A Brief Survey of Deep Learning Techniques for Person Re-identification \cite{mathurBriefSurveyDeep2020} & TCSVT20 & Described for the first time the disparity of Person Re-ID between closed and open world scenarios and pointed out the limitations of open Re-ID works.\\
    \hline
    Islam & Person search: New paradigm of person re-identification: A survey and outlook of recent works \cite{islamPersonSearchNew2020} & IVC20 & Discussed about feature representation learning and deep metric learning with novel loss functions.\\
    \hline
    Lavi et al. & Survey on Reliable Deep Learning-Based Person Re-Identification Models: Are We There Yet? \cite{laviSurveyReliableDeep2020} & arXiv20 & Investigated the SOTA DNN models now available for the topic of person re-identification and considered the restrictions of these models.\\
    \hline
    Wang et al. & Beyond Intra-modality: A Survey of Heterogeneous Person Re-identification \cite{wangIntramodalitySurveyHeterogeneous2020} & IJCAI20 & Categorized four cross-modal application scenarios in Person Re-ID: sketch, text, low resolution(LR)and infrared(IR).\\
    \hline
    Zou et al.& Person re-identification based on metric learning: a survey \cite{zouPersonReidentificationBased2021} & MTA21 & Summarized the research progress of person Re-ID methods based on metric learning.\\
    \hline
    Ye et al.& Deep Learning for Person Re-identification: A Survey and Outlook \cite{yeDeepLearningPerson2022}& TPAMI21 & Reviewed for closed-world person Re-ID from three different perspectives, including deep feature representation learning, deep metric learning and ranking optimization\\
    \hline
    Yang et al.& Survey on Unsupervised Techniques for Person Re-Identification \cite{yangSurveyUnsupervisedTechniques2021a}& CDS21 & Surveyed the SOTA unsupervised Approaches of person Re-ID.\\
    \hline
    Yaghoubi et al.& SSS-PR: A short survey of surveys in person re-identification \cite{yaghoubiSSSPRShortSurvey2021}& PRL21 & Presented a multi-dimensional taxonomy to classify the most pertinent person Re-ID studies based on diverse viewpoints.\\
    \hline
    Wang et al.& Cross-Domain Person Re-identification: A Review \cite{wangCrossDomainPersonReidentification2021}& AIC21 & Investigated the available datasets for cross-domain person re-identification and compared the results of existing methods these on the datasets.\\
    \hline
    Ming et al.& Deep learning-based person re-identification methods: A survey and outlook of recent works \cite{mingDeepLearningbasedPerson2022}& IVC22 & Suggested to divide deep learning based person re-identification approaches to four groups, including deep metric learning, local feature learning, generative adversarial learning, and sequence feature learning methods.\\
    \hline
    Peng et al.& Deep Learning-Based Occluded Person Re-Identification: A Survey \cite{pengDeepLearningBasedOccluded2023}& TMCC22 & Proposed a deep learning -based survey of occluded person re-identification methods.\\
    \hline
    Singh et al.& A comprehensive survey on person re-identification approaches: various aspects \cite{singhComprehensiveSurveyPerson2022} & MTA22 & Discussed person re-identification approaches based on image/video data in terms of temporal, spatial, metric, and automation dimensions.\\
    \hline
    Huang et al.& Deep learning for visible-infrared cross-modality person re-identification: A comprehensive review \cite{huangDeepLearningVisibleinfrared2023}& IF22 & Provided a comprehensive classification for those SOTA visible-infrared cross-modality person re-identification models.\\
    \hline
    Zahra et al.& Person re-identification: A retrospective on domain specific open challenges and future trends \cite{zahraPersonReidentificationRetrospective2023}& PR23 & Overviewed the work on image/video-based person re-identification in four aspects: dataset, architecture, number of papers, and challenges.\\
    \hline
  \end{tabularx}
  \end{table*}

  \begin{figure*}[!t]
    \centering
    \includegraphics*[width=\textwidth]{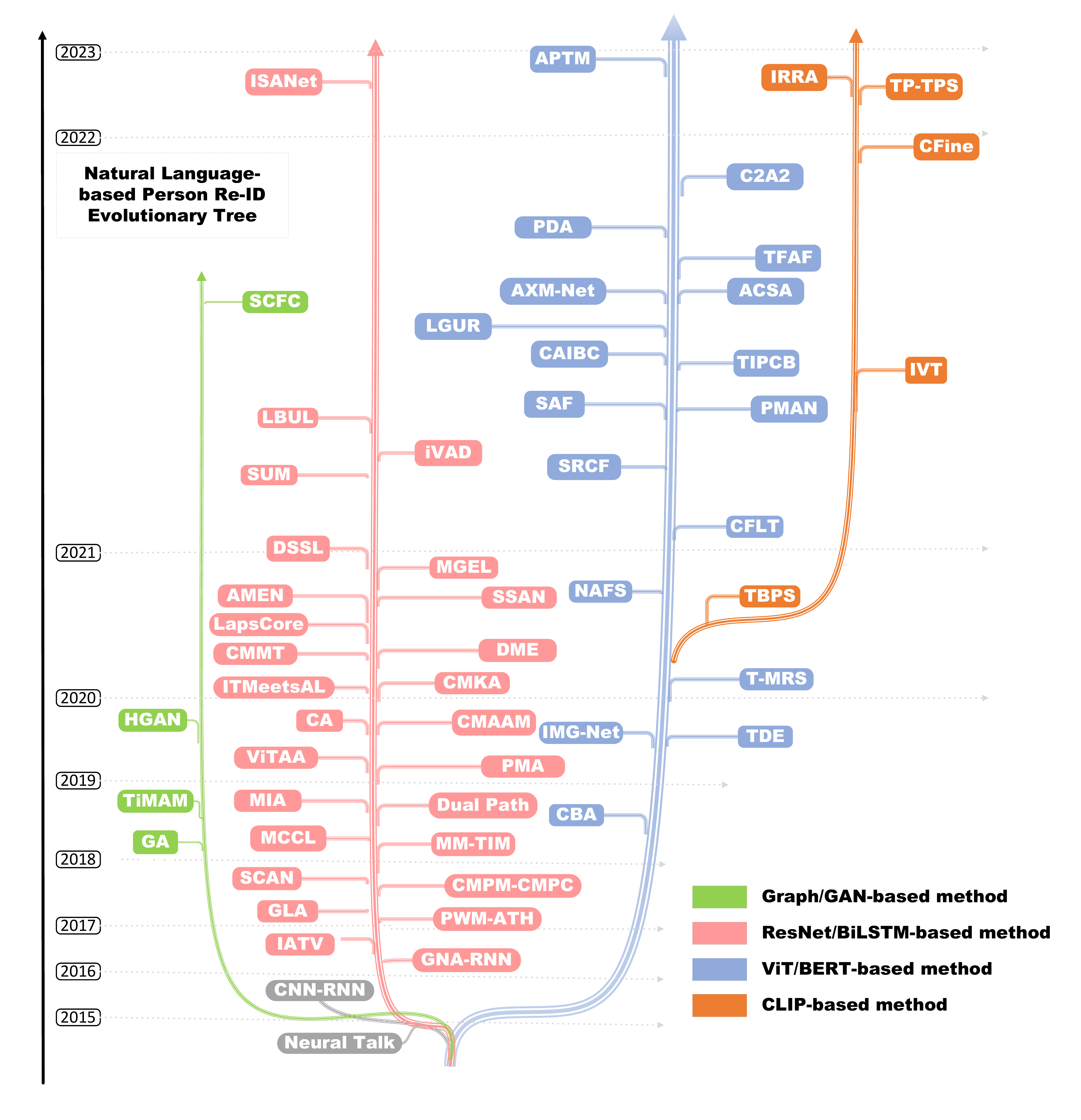}
    \caption{The figure shows a tree development diagram of all existing text-based pedestrian re-recognition methods. In the vertical direction from bottom to top represents the sequence of time development, and in the horizontal direction different trunk colors represent different trunk methods. \textcolor{green}{Green} represents Graph/GAN-based methods, \textcolor{pink}{pink} represents ResNet/BiLSTM-based methods, \textcolor{blue}{blue} represents ViT/BERT-based methods, \textcolor{orange}{orange} represents CLIP multimodal macromodel-based methods, and grey represents text-based pedestrian re-recognition research prior to its emergence, which also appears as a comparison method.}
    \label{fig3}
  \end{figure*}

  
  \begin{landscape}
      \begin{table}[htbp]
      \centering
      \caption{Text-based person Re-ID SOTA methods}
      \label{sota}
  \resizebox{700pt}{!}{
  \scriptsize
  \begin{supertabular}{|c|c|c|c|c|c|c|c|c|c|c|c|c|c|c|c|c|}
  

  \hline
   & & & & & & & & &\textbf{CUHK-PEDES}& & &\textbf{ICFG-PEDES}& & &\textbf{RSTPReid}&  \\
  \hline
  \textbf{Method} & \textbf{Publication} & \textbf{Text-to-Image} & \textbf{Feature} & \textbf{Visual backbone} & \textbf{Language backbone} & \textbf{Auxiliary strategy} & \textbf{Alignment loss} & \textbf{Rank@1} & \textbf{Rank@5} & \textbf{Rank@10} & \textbf{Rank@1} & \textbf{Rank@5} & \textbf{Rank@10} & \textbf{Rank@1} & \textbf{Rank@5} & \textbf{Rank@10} \\
  \hline



  CNN-RNN[46] & CVPR16 & TI & global & VGG-16 & Bi-LSTM & / & ID Loss & 8.07 & \ & 32.47 & & & & & & \\
  \hline
  Neural Talk[42] & CVPR15 & TI & global & VGG-16 & Bi-LSTM & / & ID loss & 13.66 & \ & 41.72 & & & & & & \\
  \hline
  GNA-RNN[4] & CVPR17 & TI & global & VGG-16 & Bi-LSTM & Attention & ID Loss & 19.05 & \ & 53.64 & & & & & & \\
  \hline
  IATV[17] & ICCV17 & TI & global & VGG-16 & LSTM & Spatial Attention & CMCE loss & 25.94 & \ & 60.48 & & & & & & \\
  \hline
  PWM-ATH[20] & WACV18 & TI & global & VGG-16 & Bi-LSTM & Attention & ID Loss & 27.14 & 49.45 & 61.02 & & & & & & \\
  \hline
  GLA[16] & ECCV18 & TI & Multi-scale & ResNet-50 & Bi-LSTM & Parts, Mask & ID loss & 43.58 & 66.93 & 76.26 & & & & & & \\
  \hline
  CMPM-CMPC[18] & ECCV18 & IT, TI & global & MobileNet & Bi-LSTM & / & CMPM-CMPC loss & 49.37 & 71.69 & 79.27 & 43.51 & 65.44 & & & & \\
  \hline
  SCAN[47] & ECCV18 & TI, IT & Multi-scale & ResNet-101, Faster R-CNN & Bi-GRU & Region/Attention & Triplet loss & 55.86 & 75.97 & 83.69 & 50.05 & 69.65 & & & & \\
  \hline
  MCCL[48] & ICASSP19 & TI & global & MobileNet & Bi-LSTM & Text map Attention & KL divergence, Triplet loss & 50.58 & \ & 79.06 & & & & & & \\
  \hline
  A-GANet[22] & MM19 & TI & Local & ResNet50/GAC & Bi-LSTM/GAC & Region, GAN, GCN & CMPM loss & 53.14 & 74.03 & 81.95 & & & & & & \\
  \hline
  TiMAM[21] & ICCV19 & IT, TI & global & ResNet-101 & BERT/LSTM & GAN & binary cross-entropy loss & 54.51 & 77.56 & 84.78 & & & & & & \\
  \hline
  CBA[49] & ICCVW19 & TI & global & ResNet-101 & Bi-GRU & Attention & ID loss, triplet Loss & 57.84 & 78.33 & 85.43 & & & & & & \\
  \hline
  Dual Path[50] & TOMM20 & TI & global & ResNet-50 & ResNet-50 & Data distribution & Instance loss, ranking loss & 44.40 & 66.26 & 75.07 & 38.99 & 59.44 & & & & \\
  \hline
  MIA[23] & TIP20 & TI & Multi-scale & VGG-16 & Bi-LSTM & Region & binary cross-entropy loss & 53.10 & 75.00 & 82.90 & 46.49 & 67.14 & & & & \\
  \hline
  GALM PMA[51] & AAAI20 & TI & global & ResNet-50 & Bi-LSTM & Pose Estimator & ID loss, Rank loss & 53.81 & 73.54 & 81.23 & & & & & & \\
  \hline
  TDE[52] & MM20 & TI & global & ResNet-101 & BERT & Attention & CMPM loss & 55.25 & 77.46 & 84.56 & & & & & & \\
  \hline
  ViTAA[14] & ECCV20 & TI & Multi-scale & ResNet-50 & Bi-LSTM & Attribute & contrastive loss & 55.97 & 75.84 & 83.52 & 50.98 & 68.79 & 75.78 & & & \\
  \hline
  IMG-Net[53] & JEI20 & TI & Multi-scale & ResNet-50 & BERT & parts & ID loss, triplet loss & 56.48 & 76.89 & 85.01 & & & & 37.60 & 61.15 & 73.55 \\
  \hline
  CMAAM[54] & WACV20 & TI & Multi-scale & MobileNet & Bi-LSTM & Attribute & ID loss, triplet loss, attribute loss & 56.68 & 77.18 & 84.86 & & & & & & \\
  \hline
  HGAN[15] & MM20 & IT, TI & Multi-scale & ResNet-50 & BERT & Parts, NLTK & ID loss, triplet loss, pair-wise loss & 59.00 & 79.49 & 86.60 & & & & & & \\
  \hline
  CA[55] & Sensors20 & TI & global & ResNet-50 & Bi-LSTM & cubic attention & ID loss, Ranking loss & 60.73 & 78.63 & 84.96 & & & & & & \\
  \hline
  CMKA[56] & TIP21 & TI, IT & global & ResNet50 & Bi-LSTM & / & KL divergence, ID Loss & 54.69 & 73.65 & 81.86 & & & & & & \\
  \hline
  ITMeetsAL[57] & PR21 & TI, IT & global & ResNet-152 & Bi-LSTM & Shannon theory and adversarial learning & ID/triplet loss, KL divergence & 55.72 & 76.15 & 84.26 & & & & & & \\
  \hline
  DME[58] & Neurocomuting21 & TI & Multi-scale & ResNet50 & Bi-LSTM & Self-attention, part & ID Loss, CMM loss, RSP loss, CMR loss & 56.32 & 77.23 & 84.71 & & & & & & \\
  \hline
  CMMT[59] & ICCV21 & TI & global & ResNet-50 & Bi-LSTM & Clustering & CMPM Loss & 57.10 & 78.14 & 85.23 & & & & & & \\
  \hline
  T-MRS[60] & TCSVT21 & TI & Multi-scale & ResNet-101 & BERT & Vanilla / overlapped/ keypoints slicing & ID/contrastive loss & 57.67 & 78.25 & 84.93 & & & & & & \\
  \hline
  TBPS[61] & BMVC21 & TI, IT & global & CLIP/ResNet-101 & CLIP-TE, Bi-GRU & CLIP & contrastive loss, ID loss & 64.08 & 81.73 & 88.19 & & & & & & \\
  \hline
  NAFS[62] & arXiv21 & TI & Adaptive Full-scale & ResNet-50 & BERT-Base-Uncased & Parts,contextual non-local attention & CMPM/CSAL loss & 59.94 & 79.86 & 86.70 & & & & & & \\
  \hline
  AMEN[63] & PRCV21 & TI & global & ResNet-50 & Bi-GRU & Graph, GAN, Autoencoder& ID/triplet/ranking/adversarial loss & 57.16 & 78.64 & 86.22 & & & & 38.45 & 62.40 & 73.80 \\
  \hline
  MGEL[64] & IJCAI21 & TI & Multi-scale & ResNet-50 & Bi-LSTM & Multi-head attention & ID/triplet Loss & 60.27 & 80.01 & 86.74 & & & & & & \\
  \hline
  SSAN[65] & arXiv21 & TI & Multi-scale & ResNet-50 & Bi-LSTM & Region, MASK & ID/Compound Ranking (CR) loss & 61.37 & 80.15 & 86.73 & 54.23 & 72.63 & 79.53 & 43.50 & 67.80 & 77.15 \\
  \hline
  DSSL[24] & MM21 & TI & global & ResNet-50 & Bi-GRU & Attention & ID/Alignment/MEC loss & 59.98 & 80.41 & 87.56 & & & & 39.05 & 62.60 & 73.95 \\
  \hline
  CFLT[66] & TIP22 & TI & Multi-scale & ResNet-50 & BERT & RVCF, Transformer & CMPM loss & 60.10 & 79.60 & 86.34 & & & & & & \\
  \hline
  SRCF-BERT[67] & ECCV22 & TI & Multi-scale & ResNet-50 & BERT & Denoising filter & ID/Compound Ranking (CR)/SEP loss & 64.04 & 82.99 & 88.81 & 57.18 & 75.01 & 81.49 & & & \\
  \hline
  iVAD[68] & Neurocomuting22 & TI & Multi-scale & VGG-16 & Bi-LSTM & Virtual Attributes decoupling & CMC/clc/ipe loss & 58.35 & 78.22 & 84.84 & & & & & & \\
  \hline
  SUM[69] & KBS22 & TI & global & ResNet-50 & Bi-GRU & Attention & ID loss, Ranking loss & 59.22 & 80.35 & 87.60 & & & & 41.38 & 67.48 & 76.48 \\
  \hline
  SAF[70] & ICASSP22 & TI & global & ViT-Base & BERT & Multi-head attention & CMPM loss, KL & 64.13 & 82.62 & 88.40 & & & & & & \\
  \hline
  PMAN[71] & PRCV22 & TI & Multi-scale & ResNet-50 & BERTBase-Uncase & multi-scale attention & CMPM loss & 64.51 & 83.14 & 89.15 & & & & & & \\
  \hline
  IVT[72] & ECCV22 & TI & Multi-scale (Text) & ViT-Base & Transformer & Mask+CLIP+NLTK & CMPM Loss & 65.59 & 83.11 & 89.21 & 56.04 & 73.60 & 80.22 & 46.70 & 70.00 & 78.80 \\
  \hline
  LBUL[73] & MM22 & TI & global & ResNet-50 & Bi-GRU & / & Ranking loss, ID loss & 61.95 & 81.16 & 87.19 & & & & 43.35 & 66.85 & 76.50 \\
  \hline
  TIPCB[5] & Neurocomuting22 & TI & global & ResNet-50 & BERT & mask & CMPM loss & 63.63 & 82.82 & 89.10 & 54.96 & 74.72 & 81.89 & & & \\
  \hline
  CAIBC[74] & MM22 & TI & global & ResNet-50 & BERT & Greyscale, color mask & triplet/ranking/ID loss & 64.43 & 82.87 & 88.37 & 58.44 & 78.2 & 85.46 & 47.35 & 69.55 & 79.00 \\
  \hline
  LGUR[75] & MM22 & TI & global & DeiT-Small & BERT, Bi-LSTM & mask & ID/ranking loss,  & 64.21  & 81.94 & 87.93 & & & & & & \\
  \hline
  AXM-Net[76] & AAAI22 & TI & Multi-scale & ResNet-50 & BERT & Parts, Self-Attention for Tex & ID/triplet/affinity loss & 64.44 & 80.52 & 86.77 & 49.37 & 71.69 & 79.27 & 53.14 & 74.03 & 81.95 \\
  \hline
  ACSA[77] & IEEE Trans. MM22 & TI & Asymmetric Cross-scale & Swin transformer & BERT & Part & ACSA/CMPM-C loss & 63.56 & 81.40 & 87.70 & & & & 48.40 & 71.85 & 81.45 \\
  \hline
  TFAF[78] & IEEE SPL22 & TI & Multi-scale & Pyramid ViT & BERT, CNN & transformer & CMPM/Reconstraction loss & 65.69 & 84.75 & 89.93 & & & & & & \\
  \hline
  PDA[2] & IJCNN22 & TI & Global & ResNet-50 & BERT & Mask & CMPM loss & 65.26 & 84.58 & 89.98 & & & & & & \\
  \hline
  (C2A2)RR[26] & MM22 & TI & Multi-scale & ResNet-50 & BERT-BaseUncased & Attribute dictionary & ID/CMPC loss, KL divergence & 67.94 & 86.86 & 91.87 & 53.14 & 74.03 & 81.95 & 54.30 & 78.70 & 86.60 \\
  \hline
  SCFC[79] & ArXiv22 & TI & Multi-scale & ResNet-50 & BERT & Graph CNN, GAN & CMPM loss & 64.12 & 82.76 & 88.65 & & & & 45.88 & 70.45 & 81.30 \\
  \hline
  CFine[80] & arXiv22 & TI, IT & Multi-scale & CIIP-ViT & BERT & CLIP & CMPM-CMPC/triplet loss & 69.67 & 85.93 & 91.15 & 60.83 & 76.55 & 82.42 & 50.55 & 72.50 & 81.60 \\
  \hline
  ISANet[81] & arXiv23 & TI & Multi-scale & ResNet-50 & Bi-LSTM & Implicit local alignment & ID/ranking/triplet loss & 63.92 & 82.15 & 87.69 & 57.73 & 75.42 & 81.72 & & & \\
  \hline
  TP-TPS[82] & arXiv23 & TI & Local & CLIP-ViT & CLIP-Xformer & CLIP & integrity ranking/dynamic attribute prompt loss & 70.16 & 86.10 & 90.98 & 54.12 & 75.45 & 82.97 & 54.12 & 75.45 & 82.97 \\
  \hline
  IRRA[83] & CVPR23 & TI & Multi-scale & CLIP-ViT & CLIP-Xformer & CLIP, Self/Cross attention & ID/SDM/DRR loss & 73.38 & 89.90 & 93.71 & 63.46 & 80.24 & 85.82 & 60.20 & 81.30 & 88.20 \\
  \hline
  RaSa[84] & IJCAI23 & TI & global & ViT-B-16 & CLIP-Xformer & ALBEF & Relation/Sensitivity-Aware/contrastive loss & 76.51 & 90.29 & 94.25 & 65.28 & 80.40 & 85.12 & 66.90 & 86.50 & 91.25 \\
  \hline
  APTM[85] & MM23 & TI & global & Swin Transformer & BERT & Attribute, Mask & ID/Contrastive loss & 76.53 & 90.04 & 94.15 & 68.51 & 82.99 & 87.56 & 67.50 & 85.70 & 91.45 \\
  \hline
\end{supertabular}
}
\end{table}
\end{landscape}
 

  \begin{figure*}[!t]
    \centering
    \includegraphics*[width=\textwidth]{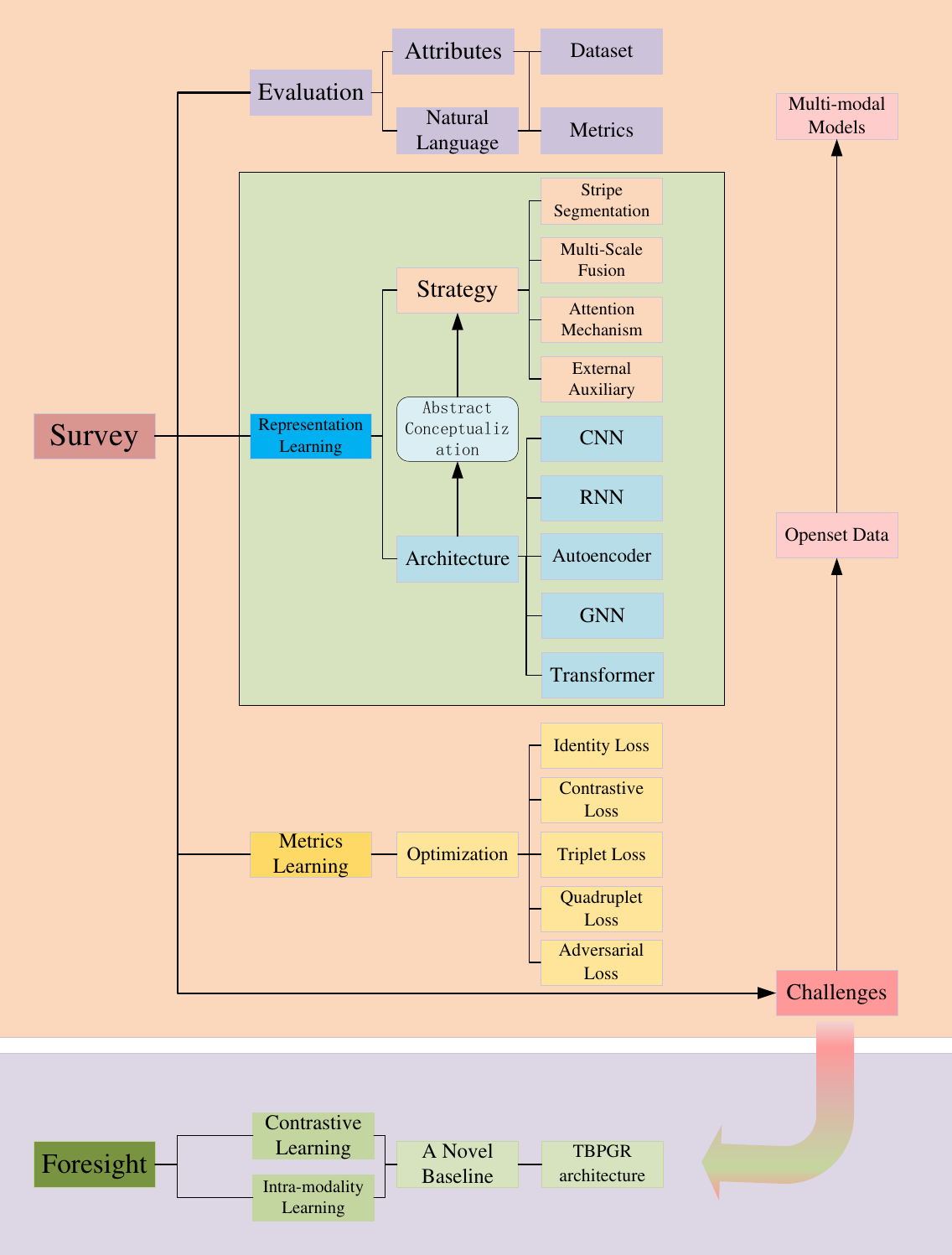}
    \caption{The figure shows the overall idea and organization regarding the running of the article, which is divided into two parts: survey and foresight. The survey part elucidates the current research on text-based pedestrian re-identification and presents some of the existing challenges. For the open-set retrieval problem, a novel baseline architecture is proposed in the foresight part.}
    \label{fig7}
  \end{figure*}

 




\vfill

\end{document}